\documentclass[journal]{IEEEtran}
\usepackage{amsmath,amssymb,epsfig,xcolor} % easier math formulae, align, subequation
\usepackage{tabularx,multirow,array}
\usepackage{times}
\usepackage{booktabs}
\usepackage{multirow}
\usepackage{soul}
\usepackage{enumerate}
\usepackage[font=small]{caption}
\usepackage[countmax]{subfloat}
\usepackage{algpseudocode}
\usepackage{algorithm}
\usepackage{caption}
\usepackage{subcaption}
\usepackage{threeparttable}
\usepackage{pifont}
\usepackage{cite}
\usepackage{url}
\usepackage{gensymb}
\usepackage{todonotes}
\newcommand{\var}{\operatorname{var}}
\usepackage[]{changes} % use final for removing text completely-- \usepackage[final]{changes}
\newcommand{\rem}[1]{}
\usepackage{setspace}

\begin{document}

\title{Leveraging Graph and Deep Learning Uncertainties to Detect Anomalous Trajectories}
% Leveraging Evidential Deep Learning Uncertainties with Graph-based Clustering to Detect Anomalies

 \author{{Sandeep Kumar Singh, \emph{Member, IEEE}, Jaya Shradha Fowdur,   Jakob Gawlikowski and Daniel Medina, \emph{Member, IEEE}}
 \thanks{\IEEEauthorblockA{\IEEEauthorrefmark{1}Sandeep Kumar Singh was with the German Aerospace Center, and he is now at the University of California, Davis, 95616, USA.}
 
 \IEEEauthorblockA{ Jaya Shradha Fowdur and Daniel Medina are with Institute of Communications and Navigation, German Aerospace Centre (DLR), Neustrelitz, 17235, Germany (e-mail: \{jaya.fowdur, daniel.ariasmedina\}@dlr.de).}
 
 \IEEEauthorblockA{ Jakob Gawlikowski is with Institute Of Data Science, German Aerospace Centre (DLR), Jena, 07745, Germany (e-mail: jakob.gawlikowski@dlr.de).} \newline
 \IEEEauthorrefmark{1}Corresponding author,  email: sansingh@ucdavis.edu
 }}
% The authors are with the Department of Nautical systems, Institute of Communication and Naviagtion,

\maketitle

\begin{abstract}
Understanding and representing traffic patterns are key to detecting anomalous trajectories in the  transportation domain. However, some trajectories can exhibit heterogeneous maneuvering characteristics despite confining to normal patterns. Thus, we propose a novel graph-based trajectory representation and association scheme for extraction and confederation of traffic movement patterns, such that data patterns and uncertainty can be learned by deep learning (DL) models. This paper proposes the usage of a recurrent neural network (RNN)-based
evidential regression model, which can predict trajectory at future
timesteps as well as estimate the data and model uncertainties associated, to detect maritime anomalous trajectories, such as unusual vessel maneuvering, using automatic identification system (AIS) data. Furthermore, we utilize evidential deep learning classifiers to detect unusual turns of vessels and the loss of transmitted signal using predicted class probabilities with associated uncertainties. Our experimental results suggest that the graphical representation of traffic patterns improves the  ability of the DL models, such as evidential and Monte Carlo dropout, to learn the temporal-spatial correlation of data and associated uncertainties. Using different datasets and experiments, we demonstrate that the estimated prediction uncertainty yields fundamental information for the detection of traffic anomalies in the maritime and, possibly in other domains.
\end{abstract}


%Nevertheless, our graphical representation of trajectory and uncertainty-based method can be applied to other transportation domains.

% \section*{\textcolor{red}{Action list}}
% \textcolor{red}{
% \begin{itemize}
%     \item \textbf{all:} end-to-end review,if not done, but focus on Sec. IV and V
% \end{itemize}
% }

 \section{Introduction}
 % Motivation for maritime shipping relevance
 \par \IEEEPARstart{S}{ecurity} is crucial for  the transportation domain. As more than 90\% of the world-trade is being carried out at sea, maritime domain in particular is facing significant challenges from natural and unnatural phenomena, such as tough environment,  collision,  illegal fishing, smuggling, pollution, and piracy \cite{UNCTAD}. Some of these challenges can be addressed by having a better understanding of maritime traffic patterns in terms of its representation, accurate localization, and its forecasting approaches. Nowadays most ships are equipped with automatic identification system (AIS) transponders to provide their static and dynamic information \cite{ais}. The AIS transponders use global navigation satellite systems for positioning, with accuracy of up to a few meters  \cite{medina2019gnss}. The vessels' location, navigational status, and voyage-related information can be used for collision-avoidance mechanisms, vessel tracking, along with the detection of loss of AIS signal, and anomalous trajectories.
 
 % Issues related with AIS misuse
 Despite being equipped with on-board transponders, some vessels meander to engage in suspicious activities. A vessel can  switch-off its AIS transponder, deviate from normal trajectory and turn in an unusual manner.  Such situations may arise, for instance, during hijacking of ships by pirates, illegal fishing or rough environment amongst others. Therefore, in addition to correctly representing traffic patterns, it is necessary to detect anomalous (or unusual) vessel trajectories, the loss of signal, unusual turns and other anomalies. As there are different kinds of anomalies in the maritime domain, they need special attention and mechanisms to be dealt with. In this context, historical and real-time AIS data  provide a valuable source of vessel trajectories information, and it has been utilized extensively for anomaly detection in the maritime domain using various methods, such rule-based, statistical  and machine learning (ML) \cite{kazemi2013open,tu2017exploiting,riveiro2018maritime,Hruschka2007, guerriero2010analysis, kowalska2012maritime, perera2012maritime, mazzarella2015knowledge,mazzarella2017novel,nguyen2021geotracknet,yuan2019novel,zhong2019vessel, singh2020machine,singh2020on}.

% Intro to how to detect anomalies in the maritime domain, types of approaches (statistical and DL)

Rule-based approaches are simple yet limited by their rigidness \cite{kazemi2013open}. In contrast, statistical and ML algorithms provide promising and complementary solutions to deal with maritime security challenges \cite{tu2017exploiting,riveiro2018maritime}. 
% Intro to statistical methods for detection
Statistical methods, such as Bayesian filtering, Gaussian processes, or  Markov models, have been proven to be very effective in the traffic movement analysis and potentially useful for anomaly detection \cite{Hruschka2007, guerriero2010analysis, kowalska2012maritime, perera2012maritime, mazzarella2015knowledge}. They are generally robust but do not exploit the short-/long-term inter-dependency of features, e.g., positional and navigation AIS information. Contrarily, the use of ML techniques, such as neural networks, long short-term memory (LSTM), have demonstrated promising results for the anomaly detection problem \cite{mazzarella2017novel,nguyen2021geotracknet,yuan2019novel,zhong2019vessel,singh2020machine,singh2020on,bensalem2019detecting,peng2021streaming,hu2006system, piciarelli2008trajectory,wang2020anomalous}. 

% Intro to DL methods for detection
ML-models are often not considered as robust under black swan events (completely unseen scenarios) due to the lack of a model confidence and an uncertainty estimation. Uncertainty estimation could help dealing with critical applications, such as anomaly detection, surveillance, and maritime navigation.  We are interested in such applications of ML techniques that include uncertainty characterization. With DL approaches \cite{gal2015bayesian,kendall2017uncertainties,sensoy2018evidential,amini2020deep}, uncertainties associated with data and models could be applied to detect anomalous trajectories and other anomalies. To date and to the authors' best knowledge, anomaly detection problem has not been approached with DL-based uncertainty estimation in maritime domain using AIS data.

% Proposal
This work leverages graph and DL-based uncertainties to deal with  maritime anomaly detection using AIS data. First, we propose a graph-based traffic representation and association (GTRA) model for extraction of maritime traffic patterns and association of new data to the patterns. The clustered data is given as input to an RNN-based regression model, which is integrated with a DL uncertainty estimator, to predict the localization and dynamics information of vessels, as well as uncertainties in their predictions. Second, we propose two different EDL classifiers to learn an unusual turn (i.e., fast maneuvers over a short time interval) and AIS on-off switching (OOS) anomalies. This article demonstrates the performance of the RNN regression models for anomalous trajectory (AT) detection, and the EDL classification models for the detection of unusual turns  (UT) and loss of AIS signal (i.e, OOS) over big and small AIS historical datasets collected from the Baltic sea region. The trained EDL classifiers are also shown to be useful in detecting the OOS and the unusual turn anomalies on a test vessel's trajectories in a validation campaign in another region, namely Bremerhaven (Germany). 

% Structure
 \par The rest of the article is organized as follows.  We present the related work in Section \ref{sec: relatedwork}. Section \ref{sec:models} describes our proposed GTRA, regression and classification models useful in detecting various anomalies.  Section \ref{sec:anomaly_defs} formulates  anomalies and detection criteria. We evaluate the performance of the models in Section \ref{sec:evaluation}, and conclude the paper in Section \ref{sec:conclusion}.

 \section{Related Work} \label{sec: relatedwork}
Detecting anomalies is a rather difficult problem as these are rare phenomena and a reference (or ground truth) dataset is generally unavailable. 
Traditionally, the anomalous trajectory detection problem has been dealt by learning and representing motion patterns together with an outlier detection criterion. Previous studies utilized clustering and graph-based approaches for traffic pattern characterization, and anomaly or similarity score for outlier detection across multiple domains and applications \cite{stauffer2000learning,hu2004survey,belhadi2020trajectory}, including social networking event mapping \cite{peng2021streaming}, tracking of automobiles  and anomalous trajectories  \cite{hu2006system, piciarelli2008trajectory,wang2020anomalous}, and extracting maritime traffic patterns \cite{pallotta2013vessel,arguedas2015spatio,xiao2017maritime,prasad2017video}. Numerous studies have used unsupervised learning methods, such as $k$-means, Gaussian mixture model, density-based spatial clustering of applications with noise (DBSCAN), etc., for traffic clustering in maritime domain \cite{pallotta2013vessel,xiao2017maritime,capobianco2021deep,coscia2018multiple, d2018detecting}.  Graph-based traffic characterization together with a similarity model is a promising approach in detecting anomalous trajectories \cite{pallotta2013vessel,karatacs2021trajectory,coscia2018multiple}. Note that the graph approaches from other domains \cite{peng2021streaming,wang2020anomalous} may not be applied directly to the maritime domain, as the nodes of a graph need to represent clusters of waypoints, i.e., areas where ships are more likely to change course, and edges should correspond to sea-lanes. Although hyper-parameters in these methods and in our GTRA are required to be tuned-- either by trial-and-error or automatically \cite{karami2014choosing}, they are able to represent motion patterns and can be useful in anomaly detection through similarity models. Nevertheless, they might not detect some complex anomalous trajectories that lie within sea-lanes, as shown in Sec. \ref{sec:evaluation}. In contrast, our GTRA method not only characterizes traffic, but also assigns new AIS data to the best possible traffic cluster (or sea-lanes) which can be useful in traffic pre-processing step for the prediction task and in the detection of heterogeneous anomalous trajectories.

Anomaly detection using either the deviation between a predicted trajectory and its reported (measured) one, or considering the  probabilistic characteristics of trajectories have also been exploited, %in the past 
using both statistical and ML-based solutions. Statistical methods, such as Kalman filter, particle filter, Bayesian networks or Gaussian processes (GPs)  have been used to estimate/predict the vessels' locations at next time stamp to reconstruct their trajectories and estimate probabilistic characteristics of trajectories to detect anomalies \cite{tu2017exploiting,riveiro2018maritime,xiao2019traffic,prasad2017video,thombre2020sensors,xiao2020big,aubry2020optimal}.  
ML (or deep learning, DL, as being branded lately) algorithms, such as LSTM and LSTM encoder decoder  \cite{hochreiter1997long, cho2014learning}, have been shown to be promising in sequence-to-sequence or time-series prediction in a number of applications, including trajectory prediction tasks \cite{yuan2019novel, liang2019neural,forti2020prediction, capobianco2021deep}.  In addition, \cite{nguyen2021geotracknet} applied the Variational RNN to learn the probabilistic distribution of trajectories, which help them to detect anomalous trajectories. 

There have also been efforts towards anomaly classification by means of supervised learning with labeled data. Zhong $et$ $al.$ \cite{zhong2019vessel} used a random forest algorithm to classify vessels using AIS data streams. Mazzarella $et$ $al.$ proposed a one-class support vector machine-based anomaly detection framework that takes AIS data as well as received signal strength into consideration for analyzing the AIS OOS anomaly \cite{mazzarella2017novel}. In the own previous works  \cite{singh2020machine, singh2020on} we showed that neural networks can be used to detect the AIS OOS anomaly with a higher accuracy than other traditional ML algorithms.     

Classical ML models generally do not indicate any measure of confidence or uncertainty with their predictions. With datasets across applications evolving over time and prone to varying noise levels, the lack of uncertainty measures in those ML models makes their use discouraged in safety-critical applications    such as anomaly detection. Thus,  it  is  highly desirable  to  quantify model and data  uncertainty in  a  trustworthy  manner. In this paper, we approach the maritime anomaly detection problem using a novel DL-based learning of model and data uncertainties. To this end, a promising EDL regression \cite{amini2020deep} and classification model \cite{sensoy2018evidential}  are applied. In a nutshell, EDL captures the model evidence (confidence) in support of its prediction. Note that the computationally-intensive, non-parametric  GPs \cite{ki2006gaussian} and parametric Bayesian neural networks (BNNs) \cite{gal2015bayesian,kendall2017uncertainties} and its approximation, known as \textit{MC dropout} \cite{gal2016dropout}, are also solid options for the provision of reliability measures and prediction uncertainty. Orthogonal to GPs and BNNs, EDL enables simultaneous learning of the desired regression/classification task along with data (aleatoric) and model  (epistemic) uncertainty estimation, by enforcing evidential priors. Furthermore, an EDL model can be trained to provide uncertainty measures without leveraging any out-of-distribution (or anomalous) data. Our work is partially inspired by this concept, wherein we propose an RNN-EDL model to detect maritime anomalies using the uncertainty estimation, and compare its performance with the MC dropout model \cite{gal2016dropout} and a graph-based similarity model \cite{karatacs2021trajectory}.  Specifically, this work makes the following contributions: 
\begin{enumerate}[1.]
\item A graph-based trajectory representation and association (GTRA) method;
\item DL models for model confidence and data uncertainty estimation towards AT, OOS and UT anomaly detection; 
\item Evaluation of anomalous trajectory detection among other anomalies on big and complex AIS dataset.
\end{enumerate}  

\begin{figure}[t]
\centering
\includegraphics[trim=0 300 50 0, clip, width=0.49\textwidth]{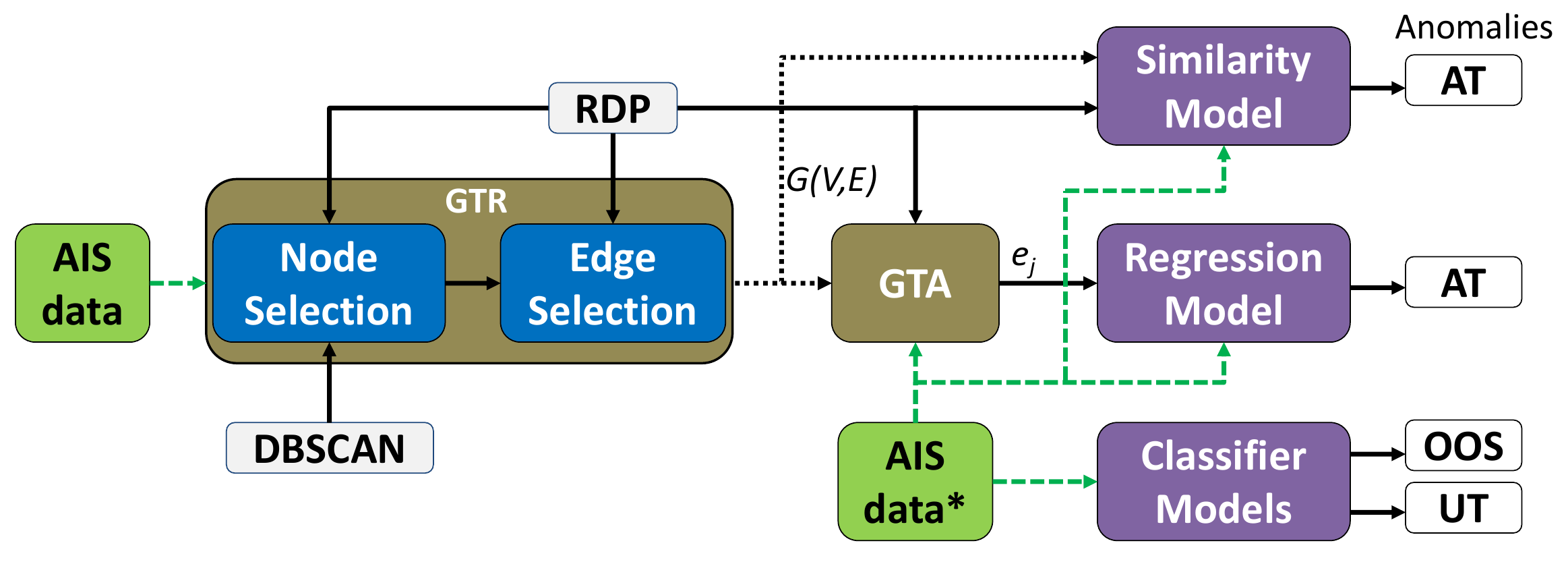}
%\vspace{-0.2 cm}
 \caption{A schematic of anomalous trajectory (AT) detection using a similarity and DL-based regression models, and OOS and UT detection using DL-based classifier models. The similarity model computes a similarity score between normal trajectory routes, obtained by a graph $G(V,E)$, and a test trajectory formed by AIS data*. $G(V,E)$ represents traffic movement patterns with nodes in $V$ and connecting edges in $E$, computed by our graph-based trajectory representation (GTR) method using DBSCAN and RDP algorithms on a given AIS dataset. A regression model tests AT on AIS data* using its trained model on AIS data and their association to the edges ($e_j \in E$) given by our graph-based trajectory association (GTA) method. Note that the classifier models use AIS data only (and no input from $G$), during their training and testing, to detect the OOS and UT anomalies.}
% \vspace{-0.4cm}
\label{fig:blockdiag}
\end{figure}

\section{Methodology for Anomaly Detection and Modeling} \label{sec:models} 
Anomaly detection is a prerequisite for the obtainment of a complete maritime traffic situational awareness. The question resides on how such anomalous patterns are defined. In general, we treat a predictable or acceptable behavior as normal and the rest is classified as anomalies. For example, the acceptable practice to find an anomalous trajectory is to examine whether the ship deviates from its well-defined maritime route. This method, also known as similarity model \cite{karatacs2021trajectory,pallotta2013vessel}, might not be suitable for vessels that may confine to maritime route but perform unusual maneuvering.  Therefore, we apply DL techniques to learn data distribution (or patterns) and their underlying uncertainties, and use them to detect anomalies. Fig. \ref{fig:blockdiag} depicts the methodology adapted in this article to detect various anomalies using our regression and classification models as well as a similarity model.

 %%\vspace{-0.3cm}
 \begin{figure}[t]
 \centering
 \includegraphics[trim=1 1 1 50, clip, width=0.4\textwidth]{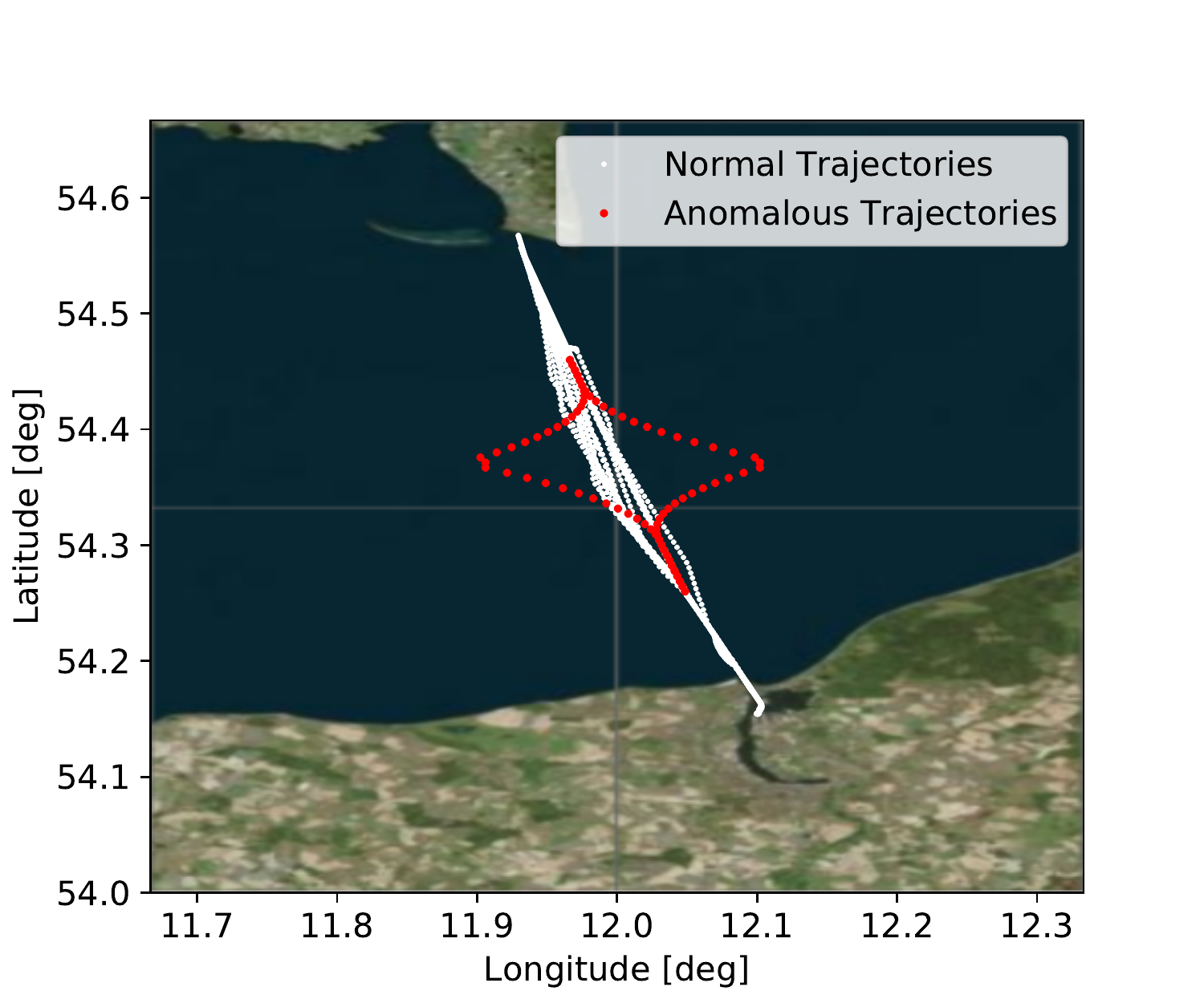}
 %\vspace{-0.2 cm}
  \caption{Normal trajectories of a vessel are shown using white dots and two anomalous segments of trajectories (left and right) are shown in red between two ports Rostock and Gedsar in the Baltic sea.}
 % \vspace{-0.4cm}
 \label{fig:rostock_gedsar}
 \end{figure}
 
One can illustrate an anomalous trajectory of a vessel with respect to its normal trajectories using an example shown in Fig. \ref{fig:rostock_gedsar}. Here, the anomalous segments  clearly deviate from the bulk of trajectories. They can be detected by either \textit{i)} thresholding the maximum allowed deviation between the posterior ground truth (reported) and the long-term a-priori prediction of a trajectory, which is not trivial for all types of vessels; or \textit{ii)} quantifying a measure of uncertainty in prediction on successive reported data. Therefore, we apply DL techniques to learn data distribution (or patterns) and their underlying uncertainties. There could be different sources of uncertainty in the prediction, when a DL model: \textit{i)} is trained with
in-distribution noisy data, \textit{ii)} is tested with mismatched or out-of-order distributed data compared to the data used in training, and \textit{iii)} has wrongly estimated weight and network parameters. The uncertainty caused by the first source can be termed as aleatoric (data) uncertainty, which is irreducible. The second and third sources cause a reducible epistemic (model) uncertainty. 
\begin{figure}[t]
\centering
\includegraphics[trim=0 320 230 0, clip, width=0.49\textwidth]{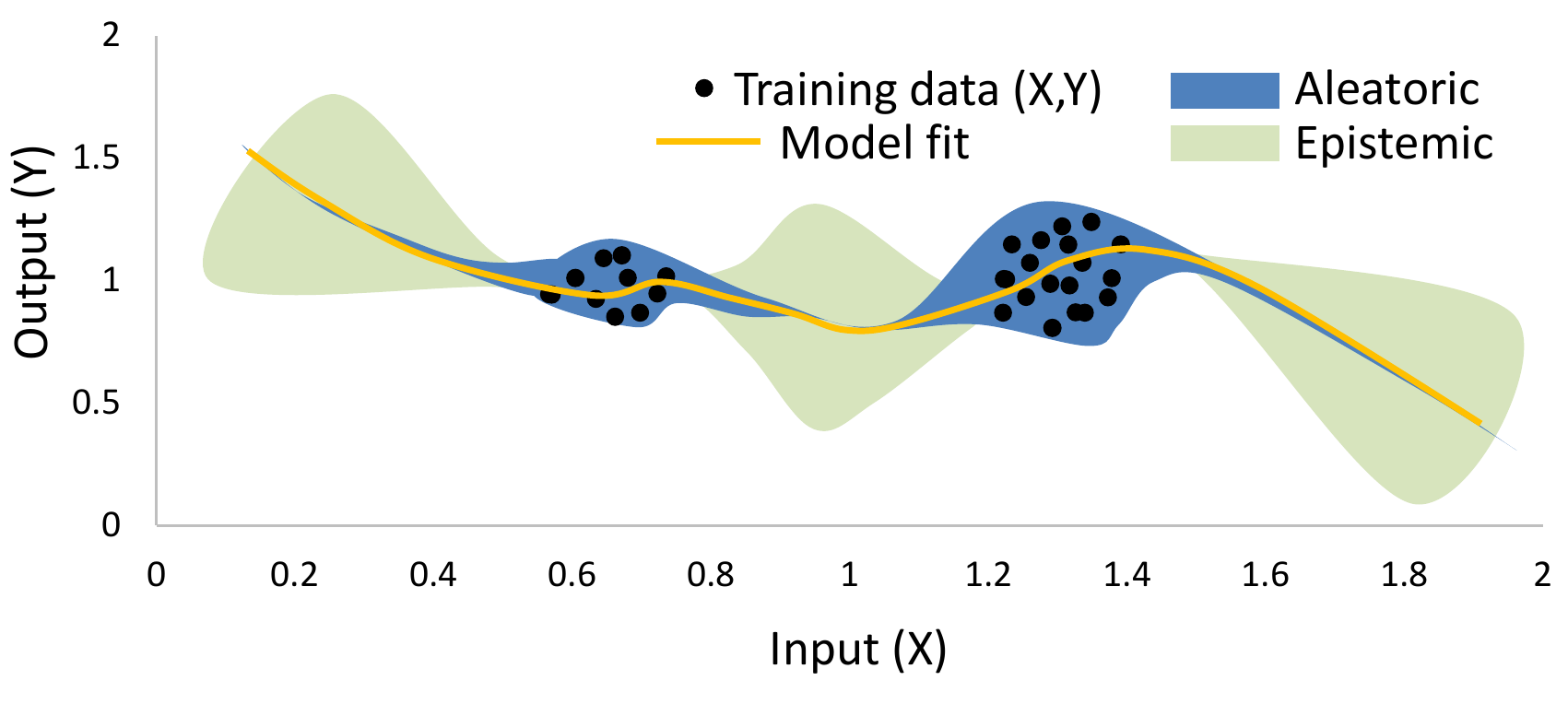}
%\vspace{-0.2 cm}
 \caption{A visualization of aleatoric and epistemic uncertainties with respect to training data and model fit.}
% \vspace{-0.4cm}
\label{fig:unc_visual}
\end{figure}
For the sake of visualization, Fig. \ref{fig:unc_visual} depicts both type of uncertainties with respect to some training data and a model-fit.
With the DL regression we could estimate the aleatoric or epistemic uncertainty for anomalous trajectory detection.% and accept/reject classification of data patterns for unusual turn, and OOS detection based on the obtained epistemic uncertainty. 

Next, we briefly describe the AIS data, and present a GTRA method, EDL regression model and EDL classification model for anomaly detection modeling. The anomalous trajectory detection problem is solved by first clustering data into well-defined routes (or sea-lanes) using the GTRA method, and then quantifying uncertainties with the clustered data using an EDL regression model.  %The GTRA model is useful in representing and associating vessels' trajectories into . %\footnote{A maritime route is defined by a well-defined set of closer tracks between two points, which is shared by many vessels at different times.}.  
%The regression model, on the other hand, forecasts the prospective positions of vessel, as well as estimate model and data uncertainties. 
%The regression model may benefit from clustered data as input. 
On the other hand, the EDL classifier models are trained to distinguish between normal and abnormal data patterns to detect unusual turns, or AIS on-off switching for instance. 

%\subsection*{Mathematical Notation \textcolor{red}{DM: to be removed later?}}Italic indicates a scalar quantity, as in $a$; lower case boldface indicates a column vector, as in $\mathbf{a}$; upper case boldface denotes a matrix, as in $\mathbf{A}$. The $n$th coordinate of the column vector $\mathbf{a}$ is denoted by $a_n$ or $[\mathbf{a}]_{n}$. The $n$th row and $m$th column element of a matrix $\mathbf{A}$ is denoted as $A_{n,m}$ or $[\mathbf{A}]_{n,m}$. The cardinality of a set $\mathcal{A}$ is denoted with $|\mathbf{A}|$. When referring to AIS-related information, the superscript indicates the vessel of interest and the subscript the time instant,  as in $\mathbf{x}^i_k$ for the information on the $i$th ship at the $k$th time instant. The sequence of AIS information for the $i$th vessel for the period between $a$ and $b$ is denoted with $\mathbf{x}^i_{a,b}$.

\subsection{AIS data representation}
AIS messages constitute the primary source of information for the methodology developed in this work. Each vessel's AIS report may include an identifier, i.e., the maritime mobile service identity ($mmsi$); kinematic information, i.e., longitude  ($lon$), latitude ($lat$), speed over ground ($sog$) and course over ground $cog$); a UTC timestamp ($time$); and other additional information, such as voyage, ship name, ship type, navigation status\textendash to name a few. %Let us denote with $\mathcal{D}$ the dataset of AIS reports used for the evaluation. 

Considering a discrete state-space model, the state vector for the navigation problem for the $i$th vessel at time index $k$ is formulated as
\begin{equation}
    \mathbf{x}_k^i \triangleq \begin{bmatrix} lon, lat, cog, sog, time, edge \end{bmatrix}^\top,
\end{equation}
with $edge$ ($e_j \in E$) being a feature that represents the association of data $\mathbf{x}_k^i$ to one of the edges of a graph $G(V, E)$ given by the GTRA scheme. It is important to note that  our anomaly detection models use different subsets of features in $\mathbf{x}_k^i$, and unlike the kinematic information directly obtained from the AIS data, the $edge$ feature is not known a priori. The notation $\mathbf{X}^i_{k:k+T}$ refers to the state of the $i$th vessel between the time periods $k$ and $k+T$, such that
\begin{equation}
    \mathbf{X}^i_{k:k+T} \triangleq \begin{bmatrix} \mathbf{x}_{k}^i, \mathbf{x}_{k+1}^i,\cdots, \mathbf{x}_{k+T-1}^i \end{bmatrix},
\end{equation}
and we use $\mathbf{X}^i$ to refer to the complete set of states for the $i$th vessel along the duration of the dataset. In the sequel, the methodology related to trajectory clustering and association, trajectory forecasting  and uncertainty estimation, and  anomaly classification are detailed.

\begin{algorithm}[t]
\caption{Graph-based Traffic Representation (GTR)}
\begin{algorithmic}[1]
\State \textbf{Input}: Dataset $\mathcal{D}$, parameters:  $\epsilon, \varepsilon, n_{min},  \mathcal{M}_{th}, e_{th}$
%\State \textbf{Output:} Graph $G$
\For {each vessel $i$ in $\mathcal{D}$}:  
	\State $\mathbf{WP}^{i} \leftarrow RDP(\mathbf{X}^i, \epsilon)$
\EndFor
%\State Nodes $V \leftarrow DBSCAN(\mathbf{WP}^i, \varepsilon, n_{min})$
\State Clusters $C \leftarrow DBSCAN(\mathbf{WP}{^i}, \varepsilon, n_{min})$
\For {each cluster $j$ in $C$}: 
    \State Nodes $V \leftarrow \mathbf{v}_j \equiv (\boldsymbol{\mu}_j, \boldsymbol{\Sigma}_j) $
\EndFor  
\Statex // define trajectory of a vessel by graph nodes  
\For {each vessel $i$ in $\mathbf{WP}$}:
    \For {each waypoint $\mathbf{wp}^i_k$ in $\mathbf{{WP}}^i$}:
            \State $\mathcal{L}^i \!\leftarrow \!\mathbf{v}_j$ if {$\min_{\mathbf{v}_j}(\mathbf{wp}^i_k\!-\!\boldsymbol{\mu}_j)' \boldsymbol{\Sigma}_j (\mathbf{wp}^i_k\!-\!\boldsymbol{\mu}_j) \!<\! \mathcal{M}_{th} $}
    \EndFor
    \State delete duplicate nodes in $\mathcal{L}^i$; delete
$\mathcal{L}_i$, if $|\mathcal{L}_i|=1$ 
\EndFor
% \For {each vessel $i$ in $\mathcal{L}$}:
%     \For{each successive pair $(\mathbf{v}_{j_1}, \mathbf{v}_{j_2})$ in $\mathcal{L}^i$}:
%         \State delete $\mathbf{wp}^i_k$ if $\mathbf{wp}^i_k==\mathbf{wp}^i_{k+1}$
%     \EndFor
%     \State delete $\mathcal{L}^i$ if $|\mathcal{L}^i| < 2$ 
% \EndFor
\Statex // compute \# of edge visits by vessels
\State $\tilde{E} := zeros(|V|, |V|)$ 
\For {each vessel $i$ in $\mathcal{L}$}:
    \For{each successive pair $(\mathbf{v}_1, \mathbf{v}_2)$ in $\mathcal{L}^i$}:
        \State $\tilde{E}(\mathbf{v}_1, \mathbf{v}_2) += 1$  // increment the visit by 1
    \EndFor
\EndFor
% \State edge\_count $\leftarrow $ sum$(e)$ \textbf{for} each edge $e$ in $\Bar{E} $
\Statex // compute the graph edges $E$
\For{each node $\mathbf{v}_1$ in $V$}:
    \For{each node $\mathbf{v}_2$ in $V$}:
        \State $E \leftarrow (\mathbf{v}_1, \mathbf{v}_2)$ if $ \frac{\tilde{E}(\mathbf{v}_1, \mathbf{v}_2)
        }{\sum \tilde{E}(\mathbf{v}_1,:)}  > e_{th}$
    \EndFor
\EndFor
\State \textbf{Return:} $G(V, E)$ for data association
\end{algorithmic}
\label{algorithm1}
\end{algorithm}

\begin{algorithm}[t]
\caption{Graph-based Traffic Association (GTA)}
\begin{algorithmic}[1]
\State \textbf{Input}: Dataset $\mathcal{D}$, graph $G$, parameters $d_{max}, \epsilon$
\For {each vessel $i$ in $\mathcal{D}$}
    \For {each time index $k$ in $\mathbf{X}^i$}
        \State $\mathbf{x}_k^i[edge] \!\!=\!\! \begin{cases}
         j, \text{if} \ \min_{1\leq j\leq |E|} d_{\bot}(\mathbf{e}_j, \mathbf{x}_k^i) \leq d_{max} \\
         \emptyset, \text{otherwise}
         \end{cases}$
	\EndFor
	\State $\mathbf{WP}^i \leftarrow RDP(\mathbf{X}^i, \epsilon)$
	\For {each $(k_1, k_2)$ time pair in $\mathbf{WP}^i$}
	    \State $ \forall \, k \in [k_1, k_2], \mathbf{E}_{k_1:k_2} \leftarrow  \{\mathbf{x}^i_k[edge]\}, $
	    \State $\mathbf{x}_k^i[edge] \leftarrow
         j, \min_{j\in \mathbf{E}_{k_1:k_2}} \vert \nabla \mathbf{e}_j - \nabla (\mathbf{x}^i_{k_1},\mathbf{x}^i_{k_2})\vert$
	\EndFor
% 	\For {each WP pair $(\mathbf{wp}^i_k, \mathbf{wp}^i_{k+1})$ in $\mathbb{WP}^i$ }
% 	     \State $\mathbb{EA}(wp_k^{k+1}) \leftarrow  \{\textbf{x}^j[edge] | \forall j \in [wp_k, wp_{k+1}]\}$ 
% 	     \State If {$|\mathbb{EA}(wp_k^{k+1})| > 1$} then 
% 	         $\forall j \in [wp_k, wp_{k+1}]: $
% 	         \State $\textbf{x}^j[edge] \leftarrow e_m \in \mathbb{EA}(wp_k^{k+1}), s.t.$ 
% 	        \State $\min_{e_m}( |slope(e_m) - slope(\textbf{x}^{wp_k}, \textbf{x}^{wp_{k+1}})|)$ \&\&  
% 	        \State $\max(\mathcal{PD}(e_m, \textbf{x}^{wp_k}), \mathcal{PD}(e_m, \textbf{x}^{wp_{k+1}})) < PD_{th}$
%     \EndFor
\EndFor
\State \textbf{Return:} $\mathbf{x}^i_k[edge], \forall i, k \in \mathcal{D}$
\end{algorithmic}
\label{algorithm2}
\end{algorithm}

\subsection{Graph-based Trajectory Representation and Association (GTRA) Scheme}
% \begin{figure}[t]
% \centering
% \includegraphics[width=0.4\textwidth]{figures/Graph-flowchart.png}
% %\vspace{-0.2 cm}
%  \caption{A flowchart to represent maritime traffic patterns through a graph and association of new AIS data to the edges of the graph. \textcolor{red}{DM: I would suggest ``encapsulating'' certain boxes and write GTR and GTA as corresponding.}}
% % \vspace{-0.4cm}
% \label{fig:graph-flowchart}
% \end{figure}
Providing an accurate traffic situation awareness requires addressing which are the usual sea lanes and vessels' trajectories. Thus, given a real-world historical AIS dataset $\mathcal{D}$, our GTRA scheme leverages on a graph-based trajectory representation (GTR) process  to conform the trajectories onto commonly-travelled maritime routes. Then, we can associate (cluster) the trajectories from any given dataset to the maritime routes via graph-based trajectory association (GTA). %Fig. \ref{fig:graph-flowchart} depicts a flowchart that summarizes our approach to represent and associate maritime traffic by a graph $G$. 
The graph $G$ consists of a set of nodes $V = \{\mathbf{v}_j\}_{j=1}^{|V|}$ and a set of edges $E = \{\mathbf{e}_j\}_{j=1}^{|E|}$, where $\mathbf{v}_j$ is a two-dimensional (2D) vector containing longitude and latitude values, and each $\mathbf{e}_j$ contains two distinct nodes in $V$. In this work, the graph's nodes represent turning points for the vessels' trajectories, while the edges joining consecutive nodes represent the sea lanes traversed by vessels. 

The GTR process, i.e., the estimation of a graph $G$, is summarized in Algorithm \ref{algorithm1}. First, in Steps 2-4 we reduce the trajectory of each vessel $\mathbf{X}^i$ to a set of waypoints $\mathbf{WP}^i$ $=\{\mathbf{wp}_j\}_{j=1}^{|\mathbf{WP}^i|}$ using an iterative Ramer-Douglas-Peucker (RDP) algorithm \cite{douglas1973algorithms}. $\mathbf{WP}^i$ may represent the trajectory's start, end and turning points, and the parameter $\epsilon$ limits the maximum distance of any intermediate points to their joining segments. The above process is repeated for all vessels' trajectories and then their associated waypoints are clustered into nodes using the DBSCAN algorithm (Step 5). DBSCAN iteratively searches for the core points in the neighbourhood of the waypoints $\mathbf{WP}^i$  within a distance of $\varepsilon$, so that there are at least $n_{min}$ number of density-reachable points. %\textcolor{red}{\st{Note that while DBSCAN  identifies an arbitrary number of clusters, it does not necessarily associate a point to a cluster, which is considered as an outlier. Then, }}
We compute the mean ($\boldsymbol{\mu}_j$) and covariance ($\boldsymbol{\Sigma}_j$) of core points in each cluster $j$ to get the nodes $\mathbf{v}_j$ in $V$ in Step 7. In Steps 9-14, the individual waypoints belonging to a vessel $i$ are then processed for proximity to the nodes in $V$. In Step 11, should the Mahalanobis distance between nodes in $V$ and a waypoint be less than threshold $\mathcal{M}_{th}$, then the closest node $\mathbf{v}_j$ for the waypoint is added to a list $\mathcal{L}^i$. Intuitively, $\mathcal{L}^i$ represents the set of nodes in $V$ that the vessel $i$ traverses. $\mathcal{M}_{th}$ represents a maximum likelihood ellipsoidal gate around node $\mathbf{v}_j$ defining points falling within the gate as those most likely to belong to the node (or trajectory). Its value was chosen based on a Chi-square distribution at a gate probability of 0.99 with 2 degrees of freedom. In addition, the nodes in every vessel's route are checked for redundancy and thereby eliminated (Step 13). Successive nodes from the resulting list $\mathcal{L}$ are then used to obtain a matrix $\tilde{E}$ that counts the number of times each edge (node-pair) is visited by vessels in Steps 15-19. Finally, in Steps 21-25 the desired edge set $E$ is obtained by applying a threshold $e_{th}$ on the normalized number of visits of each possible edge $(\mathbf{v}_1, \mathbf{v}_2)$ with respect to the total number of visits to $\mathbf{v}_1$. Thus, the GTR retains essentially the most visited edges as the trajectory (or traffic) representation. The computational complexity of the GTR is $\mathcal{O}(nm\log m + nm + n|V|+|V|^2) \approx \mathcal{O}(nm\log m)$, scaling with the number of vessels $n$, and data points per trajectory $m$, with $m >> |V|$.

The GTA process aims at associating (or clustering) all AIS data from $\mathcal{D}$ to the edges of the graph formed by the GTR, as formulated in Algorithm \ref{algorithm2}. This is performed by computing the perpendicular distances $d_{\bot}$ from an AIS data of a vessel to all edges in Steps 2-5. The edge with the minimum distance is selected if it lies below a threshold $d_{max}$, otherwise the point is associated to a so-called outlier edge  $\emptyset$. Note that this process may assign successive points of a trajectory segment to different edges. Therefore, we compute waypoints $\mathbf{WP}^i$ for each vessel's trajectory $\mathbf{X}^i$ in Step 6 using the RDP algorithm. Next, we form a set of edges that was associated (in Step 4) to all intermediate AIS data, i.e.,  $\mathbf{E}_{k_1:k_2}$ between each pair of consecutive waypoints $(\mathbf{x}^i_{k_1}, \mathbf{x}^i_{k_2}) \in \mathbf{WP}^i$ in Steps 7-8. At last, these data points are associated to an edge $\mathbf{e}_j$ that has a minimum slope  difference from a line formed by joining the waypoints (Step 9). Note that the Steps 7-11 is irrelevant if all points of a segment has been associated the same edge, i.e., $|\mathbf{E}_{k_1:k_2}| = 1$, which could also be an outlier edge. For a given dataset with $n$ vessels and $m$ data points per vessel, the computational complexity of associating all of its data points to the edges of the graph $G$ is $\mathcal{O}(nm + nm \log m) \approx\mathcal{O}(nm\log m)$.

Our intuition is that a normalcy trajectory prediction model could be learned more precisely for clustered data, as compared to raw data. Thus, in addition to AIS data features $(lon, lat, cog, sog)$, we could provide the edge label for each AIS data point as an input feature to a trajectory forecasting model, which is described next.  

\subsection{An RNN-EDL Regression Model} 
The aim of an RNN-EDL regression model is to forecast the subsequent $L$ navigation instances for an $i$th vessel, given the last $T$ time instances prior. Thus, the RNN-EDL regression model produces an estimate as
%In order to forecast the subsequent $L$ values of an $i$th vessel's trajectory based on a $k$th input sequence of length $T$, i.e., $\hat{\mathbf{X}}^i_{k+T:k+T+L} =\{\hat{\mathbf{x}}^i_j|\mathbf{X}^i_{k:k+T}, j=k+T, \ldots, k+T+L-1\}$, we introduce an RNN-EDL regression model 
\begin{equation}
    f\left(\mathbf{X}^i_{k:k+T}, \mathbf{W}\right) = \widehat{\mathbf{X}}^i_{k+T:k+T+L} , 
\end{equation}
with $\mathbf{X}^i_{k:k+T}$ the input sequence, $\mathbf{W}$ the model weights optimized during the training process and $f(\cdot)$ the estimator \textit{per se}. The model forecasts the estimates $\widehat{\mathbf{X}}^i_{k+T:k+T+L}$ for $L$ time instances.
%To that end, we employ loss functions that support  \textit{evidential regression} \cite{amini2020deep} to estimate the actual forecast and the associated (epistemic) uncertainties over the estimates. %The latter is particularly useful to perform outlier detection efficiently.
Henceforth, we will omit the vessel index $i$.

Among the recurrent networks, the LSTM encoder-decoder is known for its superior performance at temporal prediction \cite{cho2014learning} and it constitutes the learning algorithm for our work. This architecture learns sequences by first encoding an input sequence ($\mathbf{X}_{k:k+T}$) into hidden state variables ($\mathbf{h}_k = f(\mathbf{x}_k, \mathbf{h}_{k-1})$), and then maximizing the posterior log-likelihood of the estimate ($\log p(\mathbf{X}_{k+T:k+T+L}|\mathbf{h}_j, \textbf{x}_j, j\in [k,\dots,k+T-1])$). Conventional LSTM regression resembles a pointwise maximum-a-posteriori (MAP) estimation, without uncertainty measures for the estimates. However, the information on the uncertainty is as valuable as the estimate itself. To that end, we employ loss functions that support  \textit{evidential learning} \cite{amini2020deep} to estimate the actual forecast and the associated epistemic uncertainties over the estimates. Furthermore, we also implement the RNN-MC dropout model  for comparative analysis, which is relatively computationally intensive because it does not provide a close-form solution for the likelihood function, and the integration is generally computed by a Monte Carlo (MC) sampling approach \cite{gal2016dropout}. 

\begin{figure}[t]
\centering
\includegraphics[width=0.45\textwidth]{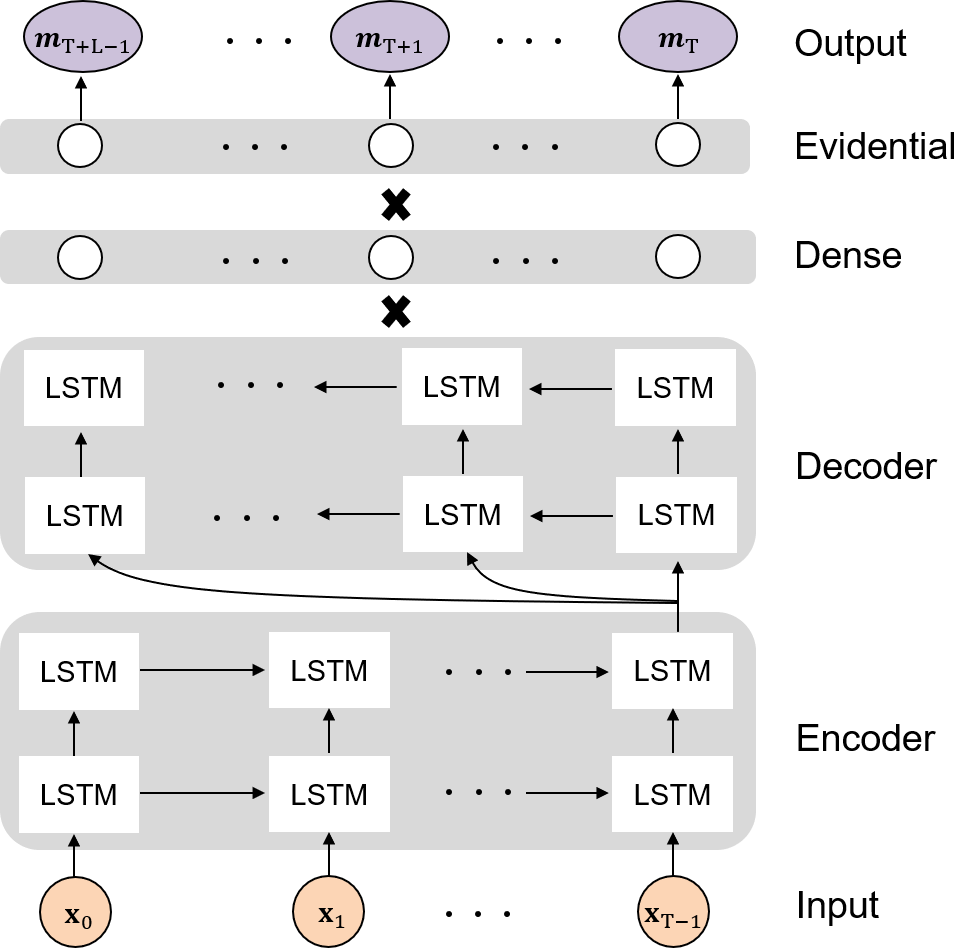}
%\vspace{-0.2 cm}
 \caption{An architecture of an LSTM Encoder-Decoder-Evidential DL model for trajectory forecasting and uncertainty estimation.}
% \vspace{-0.4cm}
\label{fig:edl_arch}
\end{figure}

The architecture of the proposed model as a trajectory predictor is depicted in Fig. \ref{fig:edl_arch}. 
The LSTM  encoder and decoder consist of a multiple stack of LSTM cells which are connected horizontally and vertically for spatio-temporal learning. The output of the LSTM decoder part is given to a Dense layer and forwarded to the Evidential layer. The Evidential layer outputs parameters $\boldsymbol{m}$ that are used to parameterize a distribution $p(x\vert \boldsymbol{m})$, based on which a MAP estimate, the aleatoric uncertainty and the epistemic uncertainty can be derived. We explain this in the following.

A basic idea of modelling data uncertainty on a prediction is to output a probability distribution $p(x\vert \boldsymbol{\theta})$ instead of a pointwise prediction \cite{kendall2017uncertainties}. For regression tasks the distribution parameters $\boldsymbol{\theta}$ might be the mean and the variance value of a normal distribution. Evidential neural networks extend this idea and model the epistemic uncertainty by a higher order distribution, i.e. a distribution over the aleatoric distribution, as
$p(\boldsymbol{\theta} \vert \boldsymbol{m})$. 
Here, $\boldsymbol{m}$ again represents the model's output which is used to parameterize a probability distribution, in our case this distribution is given by the normal inverse gamma distribution which is described in more detail below. The intuitive idea behind this is that if the model receives data which does not fit the patterns learned during training, the information that can be extracted from this data are very limited and hence the \textit{evidence} for a certain prediction (even when given as mean and variance) is also limited. As a result, the higher order distribution, which is parameterized by this evidence, represents the uncertainty as well. 

Based on the evidential modelling, the probability of a prediction $x_k$ is given by
\begin{eqnarray}
p(x_k|\boldsymbol{m})  
         = \int_{\boldsymbol{\theta}}p(x_k|\boldsymbol{\theta}) p(\boldsymbol{\theta}|\boldsymbol{m}) d\boldsymbol{\theta}.
\end{eqnarray}
For our approach, we follow \cite{amini2020deep} and model the aleatoric uncertainty by a normal distribution with unknown mean and variance, i.e.
 $\boldsymbol{\theta} = (\mu, \sigma^2)$ 
and use its conjugate prior, i.e., Normal Inverse-Gamma (NIG), as a fitting distribution for representing the epistemic uncertainty. 
The NIG distribution is parameterized by four scalar parameters, i.e.
\begin{equation}
    \boldsymbol{m}= (\hat{x}, v, \alpha, \beta) \qquad \text{with}\quad \hat{x} \in \mathbb{R},v > 0, \alpha > 1, \beta > 0.
\end{equation}
\begin{equation}
    \mu \sim \mathcal{N}(\hat{x},\sigma^2/v) \qquad \text{and}\qquad \sigma^2 \sim \Gamma^{-1}(\alpha, \beta)
\end{equation}
where $\Gamma^{-1}$ is the inverse gamma distribution.

For the probability of a prediction we have to consider the marginal likelihood $p(x_k|\boldsymbol{m})$, which is given as the Student's-t
\begin{equation} \label{eqn:likelihood}
p(x_k|\boldsymbol{m}) = St(x_k; \hat{x}_k, \frac{\beta(1+v)}{v\alpha}, 2\alpha).
\end{equation}
Please refer to \cite{amini2020deep} and their supplementary materials for further details and the derivations.

During the training process the network parameters $W$ are optimized in order to minimize the negative log-likelihood 
\begin{equation}
    -\log p\left(\int_{\boldsymbol{\theta}}p(x_k|\boldsymbol{\theta}) p(\boldsymbol{\theta}|\boldsymbol{m}) d\boldsymbol{\theta})\right).
\end{equation}

Given a $k$th input sequence $\mathbf{X}_{k:k+T}$, the loss function $\mathcal{L}_k$ for a ground-truth output sequence $\mathbf{X}_{k+T:k+T+L}$, the predicted parameters $\boldsymbol{m}_j^d$ and an estimated mean $\hat{x}_j^{d}$ can be computed as \cite{amini2020deep}
\begin{flalign}
 \mathcal{L}_k(W) \!&=\! \mathcal{L}^{NLL}_k(W) + \lambda \mathcal{L}^R_k(W)  \nonumber \\
                    \!&=\! \sum_{ j=k+T}^{k+T+L-1} \!\sum_{d=1}^n \!\!- \log p(x_j^{d}|\boldsymbol{m}_j^{d}) \!+\! 
                     \lambda |x_j^{d} \!-\!\hat{x}_j^{d}| (2v_j^{d}\!+\!\alpha_j^{d}),
\end{flalign}
where the first loss component is used for minimizing the negative log-likelihood (NLL) of an observation $x_k$ and the second one is an evidence regularizer that motivates to collect less evidence when the predictions are false with a scalar $\lambda$. The RNN-EDL model is trained to output four evidential distribution parameters for each output $x_k^d,d=1,2,\ldots,n \in \mathbf{x}_k$, i.e., $\boldsymbol{m}_k^{d}=(\hat{x}_k^{d}$, $v_k^{d}$, $\alpha_k^{d}, \beta_k^{d})$  
for an $L-$length output sequence $\mathbf{X}_{k+T:k+T+L}$. 
Given a NIG distribution and following \cite{amini2020deep}, we can compute the next timestep prediction at $j=k+T$ for each feature $d=1,2,\ldots,n$, given an input sequence $\mathbf{X}_{k:k+T}$, as $\hat{x}_j^{d}=\mathbb{E}[\mu_j^{d}]$, the aleatoric uncertainty as $v_j^{d} \times \var[\mu_j^{d}]$, 
%  \begin{equation} \label{eqn:unc_data}
%   \mathbb{E}[\sigma^2_i] = \frac{\beta_i}{\alpha_i-1}
% \end{equation}
and the epistemic uncertainty by
\begin{equation} \label{eqn:unc_model}
\var[\mu_j^{d}] = \frac{\beta_j^{d}}{v_j^{d}(\alpha_j^{d} - 1)}.
\end{equation}

\subsection{EDL Classifiers for Anomalous OOS and Turn Detection}
\begin{figure}[t]
\centering
\includegraphics[width=0.49\textwidth]{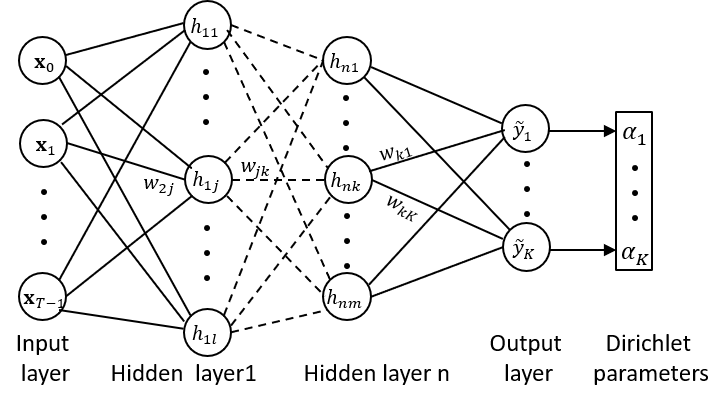}
%\vspace{-0.2 cm}
 \caption{An architecture of an EDL classifier model that consists of an input layer, $\text{n}$ hidden layers, an output layer, and a Dirichlet parameterization layer.}
% \vspace{-0.4cm}
\label{fig:edl_class}
\end{figure}
A standard neural network outputs $\boldsymbol{\tilde{y}}=f(\Theta, \mathbf{X}_{0:T}) =a_{\text{n}+1}(W_{\text{n}+1}a_\text{n}(W_{\text{n}}\ldots a_1(W_1\cdot \mathbf{X}_{k:k+T})\ldots))$ with the number of hidden layers $\text{n}$, activation functions $a_j(\cdot), j=1,2,\ldots,\text{n}$, and network parameters $\Theta = [W_{1:\text{n}+1}, \boldsymbol{b}]$, where $W$ is the weight vector of the connections between two layers, and $\boldsymbol{b}$ is a bias vector. For classification,  generally a softmax activation is used to convert the predicted values ($\tilde{y}_1, \tilde{y}_2, \ldots, \tilde{y}_K \in \mathbb{R}$) of the output layer to class probabilities $p_c = \text{softmax}(\tilde{y}_c) = \exp(\tilde{y}_c)/\sum_{j=1}^K\exp(\tilde{y}_j)$. In doing so it does not take the magnitude of the logits into account and hence fails to represent the certainty of the predicted probability vector outputs. Therefore, authors in \cite{sensoy2018evidential} proposed to replace the softmax activation with any other continuous operator, for example $ReLU(x) = \max(x, 0)$, and consider $\boldsymbol{\alpha} = f(\Theta,\mathbf{X}_{k:k+T})+1$ (i.e., output+1)  as parameters of a multivariate Dirichlet distribution $\boldsymbol{\alpha}=(\alpha_1, \alpha_2, \ldots, \alpha_K)$. An architecture of EDL classifier is depicted in Fig. \ref{fig:edl_class}. Notice that unlike the evidential regressor, which models outputs with the continuous Gaussian distribution, the evidential classifier utilizes the categorical Dirichlet's distribution, with a sum of squares loss function $||y_c - \tilde{p}_c||^2_2$. Once the network learns the parameters $\boldsymbol{\alpha}$, its mean, can be taken   
 as an estimate of the class probabilities, shown in Eq. \eqref{eqn:edl_prob}. The epistemic uncertainty on the prediction is computed as the inverse of total evidence or Dirichlet strength $S = \sum_{c=1}^K \alpha_c$, for $K$ output classes in \eqref{eqn:unc_classifier}\cite{sensoy2018evidential}.
\begin{equation}\label{eqn:edl_prob}
\tilde{p}_c = \alpha_c/S, c= 1, 2, \ldots, K
\end{equation}
\begin{equation}\label{eqn:unc_classifier}
u = K/S
\end{equation}

We train two EDL classifiers, one for the UT detection and other for the AIS OOS detection, with different inputs and networks required for the learning purposes. The prediction uncertainty in Eq. \eqref{eqn:unc_classifier} together with \eqref{eqn:edl_prob} is used to accept or reject the classification of data patterns (e.g., for UT, OOS) by setting a threshold on acceptable uncertainty $u_{th}$. In the next section, we describe how we detect different anomalies using the proposed models. 

\section{Anomaly Definitions and Detection Criteria}\label{sec:anomaly_defs}
The following subsections describes three anomalies and presents the detection criteria.    

\subsection{Anomalous Trajectory (AT)} 
For AT detection, we divide a trajectory into smaller segments, each with $N+T-1$ consecutive AIS messages, and cover it by a sliding window of length $T$ with a step size of one. This results into $N$ sequences per segment. To simplify the evaluation of uncertainties, we keep the output sequence length as $L=1$. %Note that each output predicted AIS data contains $n$ features, where two features\textendash $lon$ and $lat$\textendash are considered to evaluate the anomalous trajectories in Sec. \ref{sec:evaluation} using the criteria described below. 
%\hl{High epistemic uncertainty may represent anomalous trajectory.... Fig. 11 rescaled, check with s.d.}

A vessel's trajectory is termed as anomalous if it contains one or more anomalous segments. Note that high epistemic uncertainty may represent anomalous trajectory. However, as the spatial maritime data is noisy and cluttered, different output features per AIS data are predicted with different uncertainties and it might not be a good idea to analyze a given track with respect to the uncertainty-threshold of other normal tracks. Thus, we define a trajectory segment as anomalous if the predicted sequences of the segment have an abrupt transition in their  epistemic (or aleatoric) uncertainties. Precisely, a contiguous segment $\mathbf{X}_{0:N+T}$ with $N$ sequences is termed as anomalous if the minimum segment uncertainty over the normalized uncertainties obtained over all the predicted outputs is smaller than a threshold $\Theta_{AT}$, i.e., 
\begin{equation} \label{eqn:at_def}
\min_{d} \left[ \frac{\min_j (\var[\mu_j^{d}])}{\max_j(\var[\mu_j^{d}])}\right] < \Theta_{AT}, 
\end{equation}
 
where Eq. \eqref{eqn:at_def} detects some abrupt transitions in the output uncertainties $\var[\mu_j^{d}]$ over $j=T, T+1, \ldots, N+T-1$ sequences for each feature $d = 1,2,\dots,n$. It selects the feature $d$ and output sequence $j$ with the minimum normalized uncertainties. If this value is below $\Theta_{AT}$, then the segment is considered as anomalous. The normalization of the uncertainties of all $N$ predicted output sequences helps in two ways. First, it makes a fair comparison of uncertainties on all features. Second, the normalization of a set of values representing a larger transition leads to a wider range of scaled values, from closer to 0 to an upper limit of 1. Thus,  setting a maximum threshold $\Theta_{AT}$ in Eq. \eqref{eqn:at_def} helps us in detecting the larger transitions in the uncertainties, which leads to the detection of anomalous trajectories. 

To understand Eq. \eqref{eqn:at_def}, consider a trajectory segment with $N+T-1=39$ consecutive AIS data, where there are $N=30$ sequences, each with length $T=10$ data. The regression model iteratively predicts $30$ corresponding output sequences, each with a predicted sequence of length $L=1$. Let us assume that the epistemic uncertainties for the features $lon \, (d=1)$ and $lat \, (d=2)$ using Eq. \eqref{eqn:unc_model} and for 30 consecutive predictions are $\var[\mu_j^{d}] = [[0.01, 0.02], \ldots, [0.30, 0.31]],$ $d = 1,2;$ $j = 10,11,\ldots, 39$. The normalized values of uncertainties over all predicted sequences $j$ are simply $\var[\mu_j^{1}]/0.30, \var[\mu_j^{2}]/0.31$. The minimum over all predictions $j$ would result in $0.01/0.30$ and $0.02/0.31$ for the $lon$ and $lat$, respectively. Under these settings, the trajectory segment is detected as anomalous if the threshold $\Theta_{AT}$ is set larger than 1/30 (i.e., the minimum over all considered features).   %\Jakob{[We should discuss the metric motivation. From a theoretical point of view, epistemic uncertainty should be enough to consider, but maybe not in practice.]}

\begin{figure*}[ht!]
    \centering
    \begin{minipage}{.49\textwidth}
        \centering
        \includegraphics[width=1\linewidth]{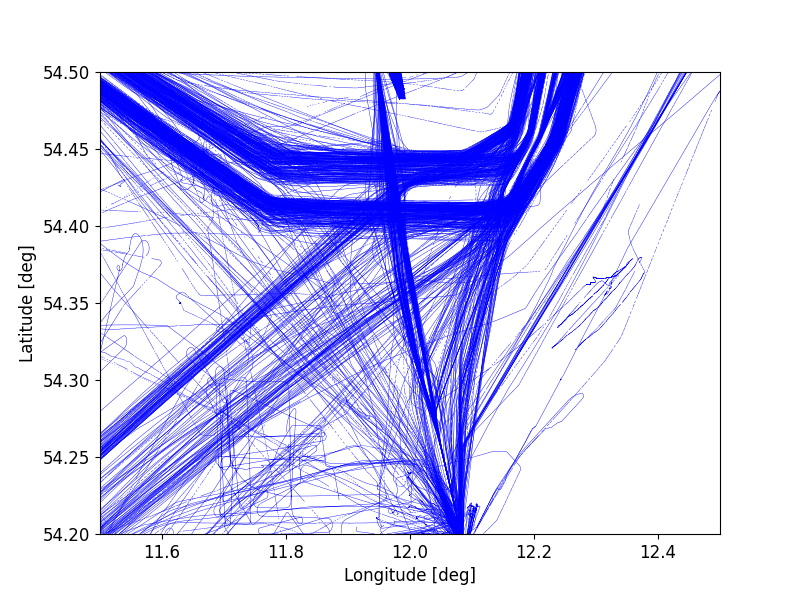}
        %\caption{Figure depicts the nodes and edges in black as created by our graph-based clustering  method using the maritime traffic data shown in blue.}
        \label{fig:dataset_2016}
    \end{minipage}
    \begin{minipage}{0.49\textwidth}
        \centering
        \includegraphics[width=1\linewidth]{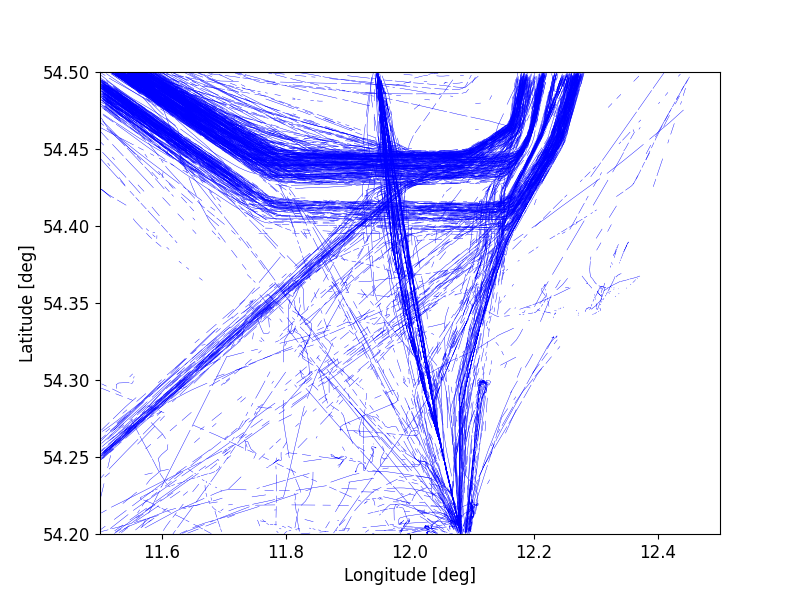}
       % \caption{Figure shows the refined nodes and edges, in black, of a graph, and the corresponding association of maritime traffic data.}
        \label{fig:dataset_2020}
    \end{minipage}
\caption{Trajectories extracted from dataset $\mathcal{D}_1$ are shown in the left plot, and trajectories from dataset $\mathcal{D}_2$ are shown in the right plot.}  
\label{fig:datasets}  
\end{figure*}

\subsection{Unusual Turn (UT) Anomaly}
We are interested in developing an EDL classifier that learns whether a vessel took a turn of more (or less) than a threshold $\Theta_{UT}$, given an AIS data sequence of length $T$. To this end, we first compute the change in the course of a vessel using the $cog \in [0\degree, 360\degree)$ values from the data sequence $\mathbf{X}_{0:T}$ by Eq. \eqref{eqn:angle_def}. 
\begin{equation} \label{eqn:angle_def}
\theta=\! 
\begin{cases}
   \max(\mathbf{M}) - \min(\mathbf{M})  +
   \max(\mathbf{L}) - \min(\mathbf{L}), \mathbf{M,L}\neq \emptyset \\
   \max(\mathbf{X}_{0:T}[cog]) - \min(\mathbf{X}_{0:T}[cog]), \text{ otherwise}
% 360\degree - \max(cog_{i:i+T-1}) + \min(cog_{i:i+T-1}) > \Theta_{UT}
\end{cases}
\end{equation}
The second line on the right hand side of Eq. \eqref{eqn:angle_def} computes the course change by the difference between the maximum and the minimum reported $cog$ values. However, the condition overlooks a scenario when a vessel turns across true north, i.e., from 359\degree to 0\degree or the other way. The first line in Eq. \eqref{eqn:angle_def} captures this scenario by formulating two sets of $cog$ values reported within  $[270\degree, 360\degree)$ using $\mathbf{M} = \{\mathbf{x}_k[cog] \in \mathbf{X}_{0:T}| \mathbf{x}_k[cog] \geq 270\degree, k=0,\ldots,T-1\}$ and within $[0\degree,  90\degree]$ using $\mathbf{L} = \{\mathbf{x}_k[cog] \in \mathbf{X}_{0:T}| \mathbf{x}_k[cog] \leq 90\degree, k=0,\dots,T-1\}$. The sum of the changes in course across the true north then correctly computes the vessels' turn  using the sample $\mathbf{X}_{0:T}$. 

A sequence $\mathbf{X}_{0:T}$ is termed as anomalous if the turn is larger than the threshold $i. e., \theta_{\mathbf{X}_{0:T}} > \Theta_{UT}$. It is important to mention that Eq. \eqref{eqn:angle_def} is used to label input data sequences into the normal and the anomaly categories, which are used for the training of the EDL unusual turn classifier. A predicted class is accepted or rejected based on its associated uncertainty.

We consider a flexible approach to define a trajectory segment for the unusual turn detection, as the duration of a vessel's turn is unknown. The segment of a track $\mathbf{X}_{k_1:k_2}$ with reported times $k_1 < k_2$ is termed as an unusual turn anomaly if all of its samples are anomalous and the previous $\mathbf{X}_{k_1-1}$ and the following $\mathbf{X}_{k_2+1}$ samples are normal. This helps in the minimization of the number of reported anomalies and the frequency that the maritime authority are to be alerted, as we noticed during a project validation campaign (results are reported in Sec. \ref{sec:evaluation}).

\subsection{AIS OOS Anomaly}
In order to detect an OOS anomaly due to loss of AIS signal, the EDL classifier model needs to learn the time difference between two consecutive AIS data. A sample $\mathbf{X}_{k:k+2}$ is termed as an AIS OOS anomaly if the time difference between two consecutive AIS data reported in the sample is greater than a threshold $\Theta_{OOS}$, as shown in Eq. \eqref{eqn:oos_def}. We either accept or reject the predicted class of a given input sample depending on whether its associated uncertainty $u$ (see Eq. \ref{eqn:unc_classifier}) is lower or higher than a threshold $u_{th}$, respectively. 
\begin{equation}\label{eqn:oos_def}
\mathbf{x}_{k+1}[time] - \mathbf{x}_k[time] > \Theta_{OOS}
\end{equation}

\section{Experimental Results}\label{sec:evaluation}
In  this  section the  proposed anomaly detection methods are evaluated on various real-world AIS data. We  describe the data preprocessing step, experimental settings, and present results from various scenarios in the following subsections. 

\subsection{Experimental Setup}
\textbf{Dataset Preparation}: The two AIS datasets evaluated were provided by: \emph{i)} the German waterways and shipping administration (WSV) for January 2016, and \emph{ii)} AISHUB for January 2020. Henceforth, they are referred to as $\mathcal{D}_1$ and $\mathcal{D}_2$, respectively. The data is filtered in the region of interest (ROI) with a rectangular bounding box from (11.5\degree, 54.2\degree) to (12.5\degree, 54.5\degree). AIS messages that reported speed over ground within 30 knots, and navigation status as 0 (i.e., underway using engine)  are considered.  Additionally, the ship type is stored (e.g., passenger: 60-69, cargo: 70-79, tanker: 80-89) for all vessels within the ROI.
After these preprocessing steps, $\mathcal{D}_1$ and $\mathcal{D}_2$ contain $\approx$ 1.2 and 0.1 million AIS positions, respectively. Figure \ref{fig:datasets} depicts trajectories of vessels that have been extracted from the datasets. Notice how the trajectories in $\mathcal{D}_2$ (shown in right) contain more missing data segments than $\mathcal{D}_1$ (shown in left). The reason is that $\mathcal{D}_1$ is satellite-based, and data in $\mathcal{D}_2$ were collected by AIS base stations.

\textbf{Models and Parameters}\footnote{The source code is at \url{www.github.com/sansastra/uncertainty_edl_graph}.}: The graph-based clustering (GTR) process in Algorithm \ref{algorithm1} uses $\epsilon=1000$ in the RDP function (Step 3), and  $\varepsilon=20$ and $n_{min}=1500$ in the DBSCAN method (Step 5). A Mahalanobis threshold $\mathcal{M}_{th} =9.2$ (Step 11) and threshold $e_{th}=0.3$ are used for the edge selection (Step 23).  The data association (GTA) process in Algorithm \ref{algorithm2}  uses a distance threshold of 7 km (Step 4),  and $\epsilon=500$ in the RDP function (Step 6). 

We use different network architectures and data feature sets for the regressor and classifiers. The RNN-EDL regressor (see Fig. \ref{fig:edl_arch}) has an encoder and a decoder, each with one LSTM layer, one Dropout layer with 0.1 dropout rate, and 128 neurons per hidden layer. The scalar constant $\lambda$ in the loss function is set to $0.01$. Moreover, each AIS data is represented by feature set $\mathbf{x}_k = \begin{bmatrix} lon, lat, cog, sog, edge \end{bmatrix}^\top$. The $edge$ feature is omitted if the model is not graph-based. An input sample consists of a $T = 10$ sequential AIS data, and output sequence length $L=1$. The data of each vessel is resampled and interpolated at $\tau = 60$ seconds. The trajectories are not interpolated for more than one hour of missing data, to deal with vessel's re-entry into ROI. For the training (50\%) and validation (10\%) of the RNN-EDL model, only cargo, passenger and tanker vessels' trajectories are used. However, the trained model is applied to test all vessel types with unseen trajectories.  %Note that each output predicted AIS data contains $n$ features, where two features\textendash $lon$ and $lat$\textendash are considered to evaluate the anomalous trajectories in Sec. \ref{sec:evaluation} using the criteria described below.
For simplicity and visualization, only two features $lon$ and $lat$ are used to compute Eq. \eqref{eqn:at_def} in the anomalous trajectory detection. Nevertheless, we also show the uncertainty variation in all, including $cog$ and $sog$, features for a single trajectory to highlight their relevance in the anomaly detection and learning process respectively.

\begin{figure*}[ht!]
    \centering
    \begin{minipage}{.49\textwidth}
        \centering
        \includegraphics[width=1\linewidth]{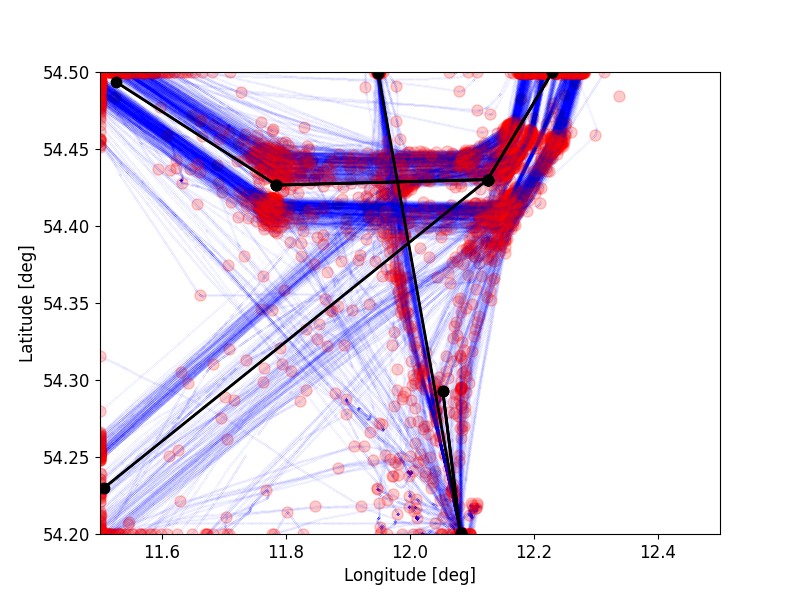}
        %\caption{Figure depicts the nodes and edges in black as created by our graph-based clustering  method using the maritime traffic data shown in blue.}
        %\label{fig:graph_original}
    \end{minipage}
    \begin{minipage}{0.49\textwidth}
        \centering
        \includegraphics[width=1\linewidth]{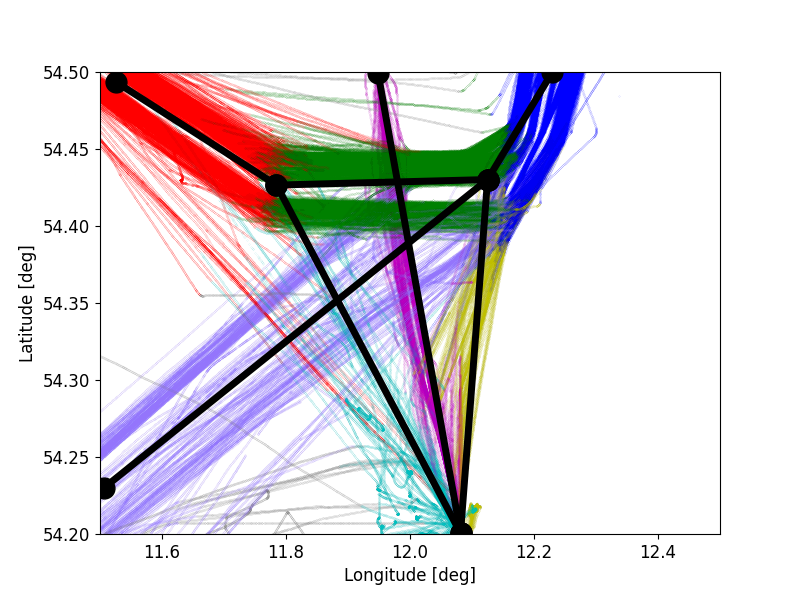}
       % \caption{Figure shows the refined nodes and edges, in black, of a graph, and the corresponding association of maritime traffic data.}
        %\label{fig:graph_refined}
    \end{minipage}
\caption{\textbf{Left} plot depicts the graph nodes and edges (in black) as created by the GTR process using the waypoints of  trajectories (in red) of maritime traffic data (in blue). \textbf{Right} plot shows the refined graph (in black), and the  association of the AIS data to one of the edges (in colors)  by the GTA process. Data shown in different colors form different clusters. An outlier cluster is shown with data points in the grey.}  
\label{fig:graph_all}  
\end{figure*}

\begin{figure*}[t]
    \centering
    \begin{minipage}{.45\textwidth}
        \centering
        \includegraphics[trim=5 0 5 30, clip,scale=0.5]{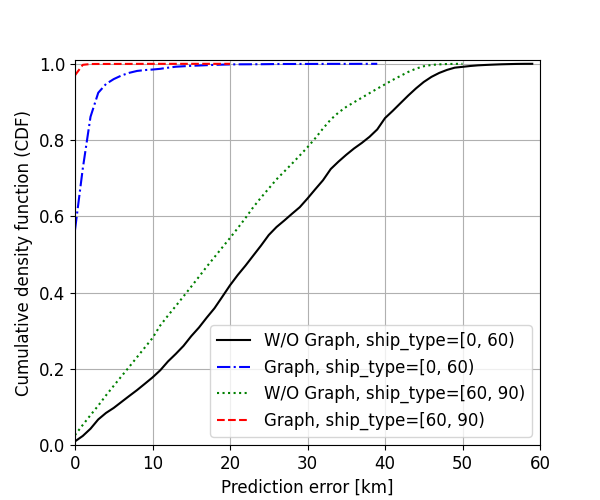}
    \end{minipage}%
    \begin{minipage}{0.45\textwidth}
        \centering
        \includegraphics[trim=5 0 5 30, clip,scale=0.5]{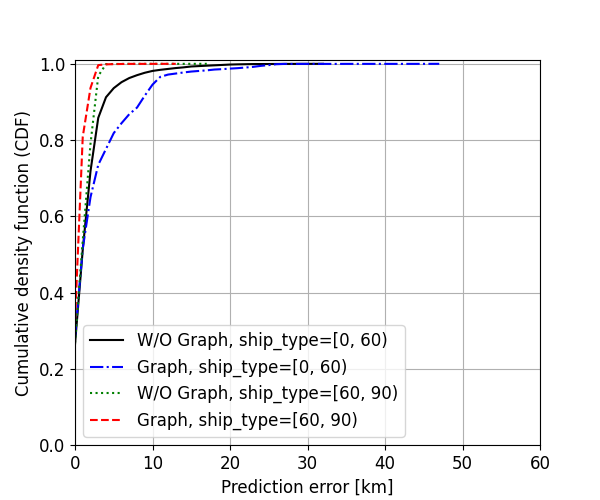}
    \end{minipage}
\caption{The plots show the CDFs on positional error between prediction  and ground truth for with and without (W/O) graph-based models, as well as for different unseen vessels of ship types [0, 60) and [60, 90) in the dataset $\mathcal{D}_1$. \textbf{Left}: Results from the RNN-EDL regression model. \textbf{Right}: Results from the RNN model without the EDL layer.}  
\label{fig:cdf_dist_error}  
\end{figure*}

The EDL classifier uses three hidden layer with 128 neurons for the unusual turn detection. An input sample has $T = 60$ consecutive AIS data with two outputs classes: normal (labeled as [1, 0]) and anomalous (labeled as [0, 1]). It uses only the $\textbf{x}_k = [cog]$ feature, and a data resampling time of $\tau=3$ seconds. The missing data in a sample is filled by linear interpolation. The EDL classifier for AIS OOS detection however, has only one single hidden layer with 128 neurons, $T = 2$ and $\textbf{x}_k = [lon, lat, cog, sog, time]$. The pre-processing of each sample assigns $\textbf{x}_k[time] \leftarrow 0$ and $\textbf{x}_{k+1}[time] \leftarrow \textbf{x}_{k+1}[time] - \textbf{x}_k[time]$, i.e., time difference between two consecutive data. Unlike the regression model, the EDL classifiers are trained (and validated) with 50\% (and 10\%) data of all ship types, while remaining 40\% data is used for testing. To make the model trained well, the number of normal and anomalous data samples are made equal, by randomly deleting some excessive normal samples in  dataset $\mathcal{D}_1$ and $\mathcal{D}_2$. All models are trained using an Adam optimiser.

\textbf{Baseline}: We use an improvised traffic routes extraction and anomaly detection method \cite{karatacs2021trajectory} as the baseline, which is a similarity model (see Fig. \ref{fig:blockdiag}) originally proposed in \cite{pallotta2013vessel}. The Similarity model uses an extracted graph $G$ to find how similar a test trajectory is as compared to the routes of a graph $\mathbb{R}(G)$. First, a test trajectory $i$ is represented by its waypoints $\mathbf{WP}^i$, for this we use the RDP with $\epsilon=1000$. Then,  it applies a Longest Common Subsequence (LCS) algorithm  \cite{hirschberg1975linear} to compute a similarity score $S_i$  between the trajectory $i$ and routes of graph $\mathbb{R}(G)$. $S_i$ is the ratio of the maximum LCS of nodes in $(\mathbb{R}(G), \mathbf{WP}^i)$ and the maximum route length. Two nodes are defined as similar if they are within $d_{max} = 7$ km range. A test trajectory is termed as anomalous if the score $S_i$ is smaller than a similarity threshold $S_{AT}$. We also implemented the RNN-MC dropout model \cite{gal2016dropout} with NLL loss function for heteroscedastic aleatoric uncertainty, keeping all the parameters same as RNN-EDL exect the EDL and its output layers. The output layer is replaced by a prediction of the mean and the standard deviation for each predicted parameter. Furthermore, the dropout is enabled during the testing phase as well and the model uncertainty is modeled as the deviation in 50 forward passes with different random dropout masks.

\textbf{Evaluation Metrics}: There is no reference dataset for anomalous tracks. A quantitative analysis in terms of accuracy or a false alarm rate is therefore not feasible. In contrast, we can evaluate the EDL classifiers for detecting OOS and UT anomalies using the accuracy measure, confusion matrix, false detection rate. Nevertheless, since anomalies are subjective in nature, each detected anomaly must undergo an inspection by data (here, AIS) experts before raising any alarm.

\begin{figure*}[t]
    \centering
    \begin{subfigure}{.45\textwidth} %\begin{minipage}{.33\textwidth}
        \centering
        \includegraphics[trim=0 0 55 45, clip,width=1\linewidth]{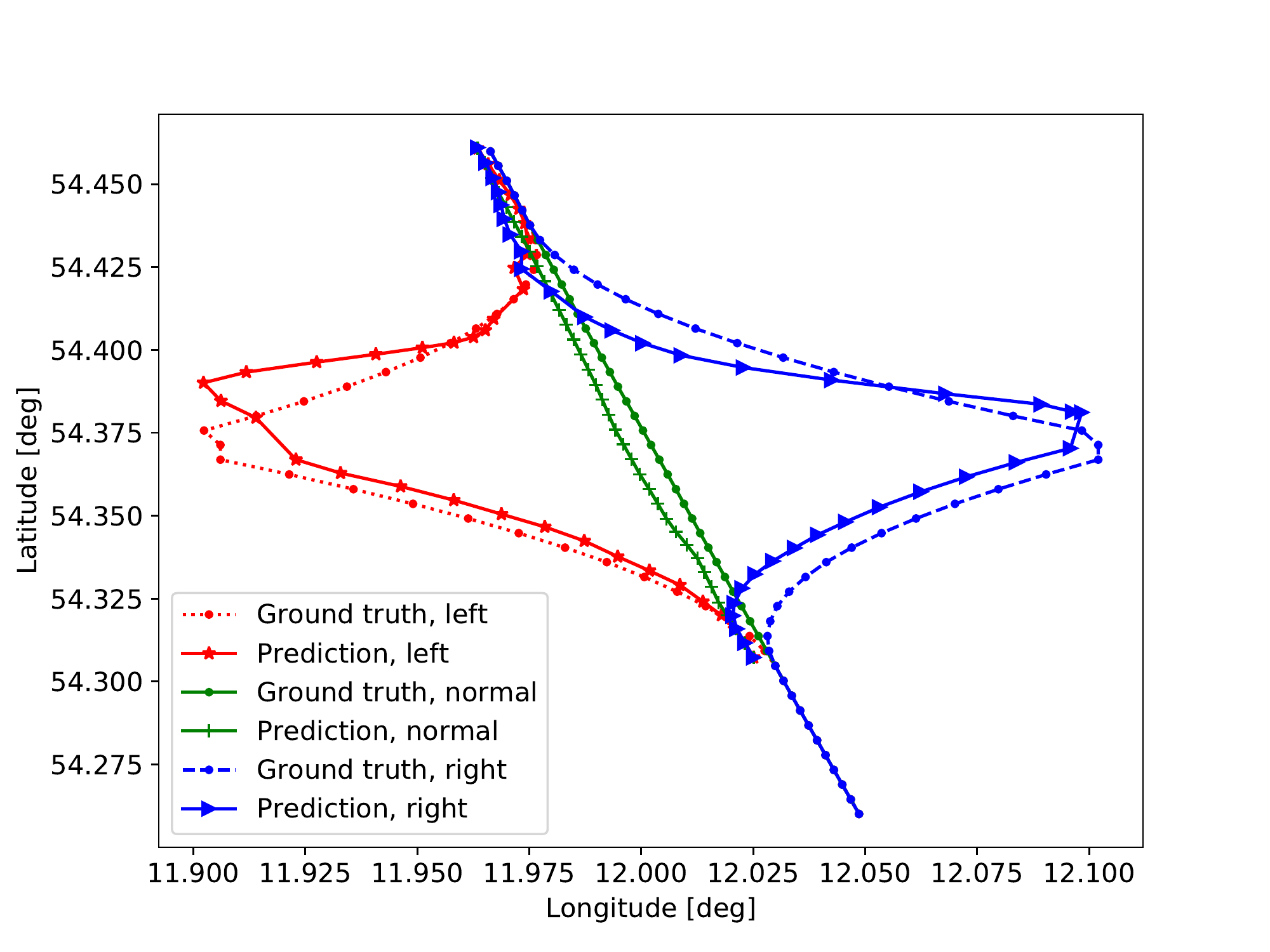}
%\vspace{-0.2 cm}
% \vspace{-0.4cm}
%\label{fig:pred_rg_left_normal_right}
    \end{subfigure} %\end{minipage}
    \begin{subfigure}{.45\textwidth} %\begin{minipage}{0.33\textwidth}
        \centering
        \includegraphics[trim=0 0 50 45, clip,width=1\linewidth]{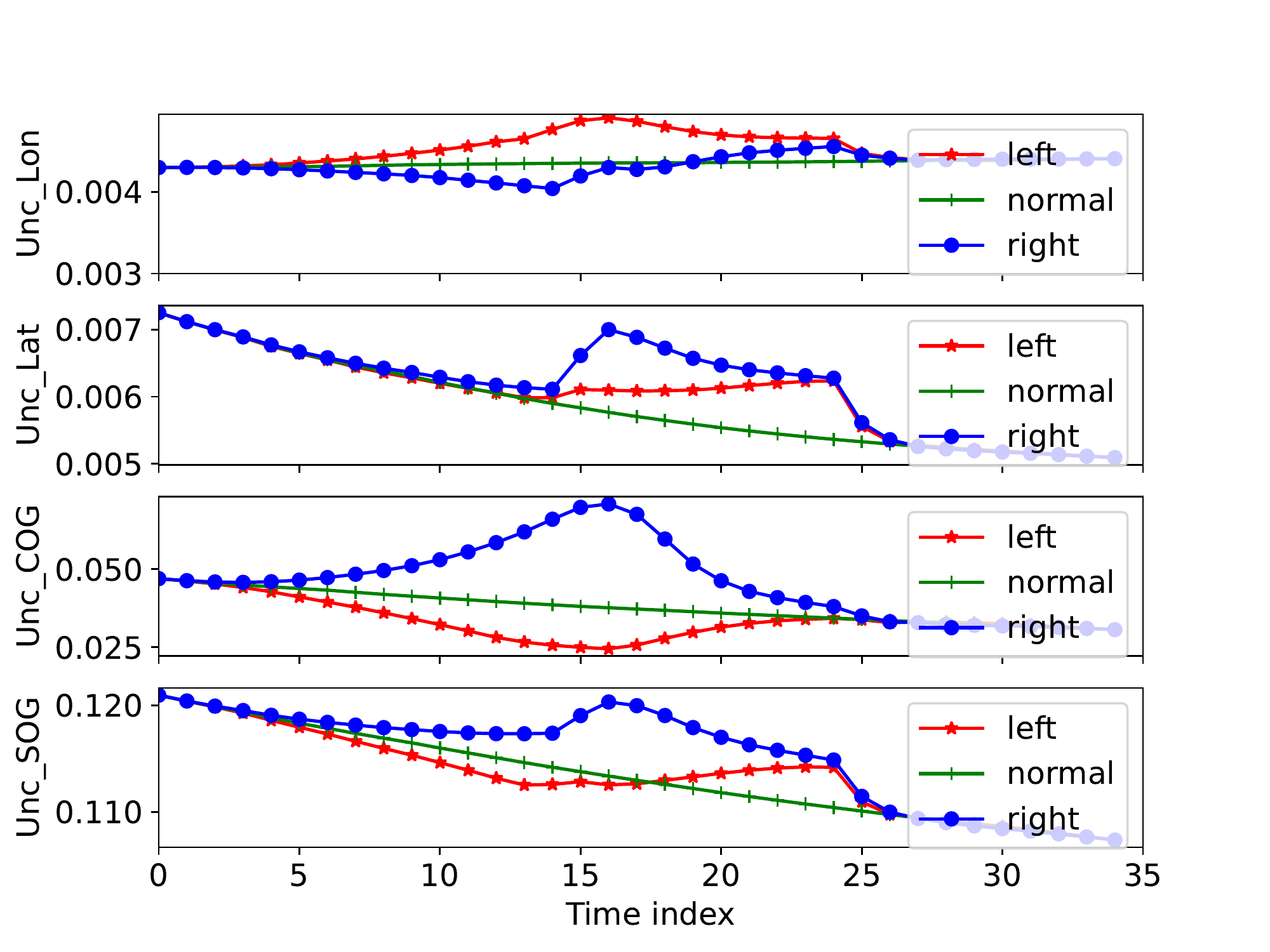}
%\vspace{-0.2 cm}
    \end{subfigure} %\end{minipage}
%  \begin{subfigure}{.32\textwidth} %\begin{minipage}{0.33\textwidth}
%         \centering
%         \includegraphics[trim=0 0 50 45, clip,width=1\linewidth]{figures/Unc_rg_all_graph_True_ds1_aleatoric.pdf}
% %\vspace{-0.2 cm}
%     \end{subfigure} %\end{minipage}   
 \caption{\textbf{Left} plot shows the ground truth and the corresponding predicted trajectories for the left, normal, and right trajectory segments by the graph-based RNN-EDL model. \textbf{Right} plots show the epistemic  uncertainties estimated on successive predictions of output data features $(lon, lat, cog, \text{and } sog)$ for the left, normal, and right trajectory segments by the model.}
% \vspace{-0.4cm}
\label{fig:pred_unc_rg_left_normal_right}
\end{figure*}

% putting percentage between two minipage minimizes gaps
\begin{figure*}[t]
    \centering
        \begin{subfigure}{.25\textwidth}
        \centering
        \includegraphics[trim=0 0 40 0, clip, width=1\linewidth]{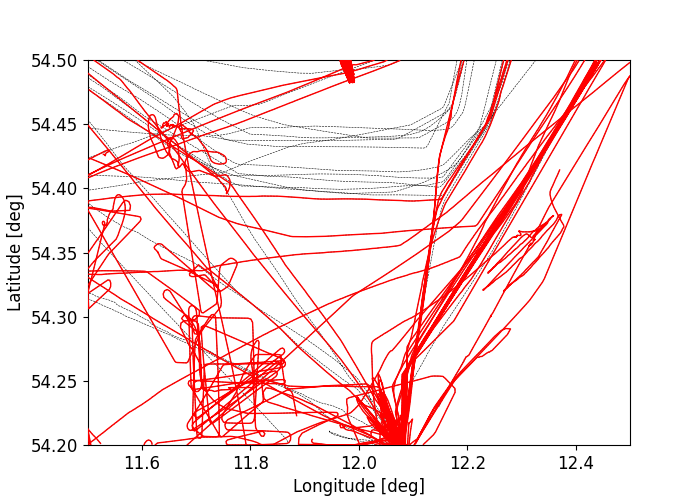}
        \caption{ship type=[0,60), $S_{AT}$=0.3}
    \end{subfigure}%
    \begin{subfigure}{0.25\textwidth}
        \centering
        \includegraphics[trim=0 0 40 0, clip, width=1\linewidth]{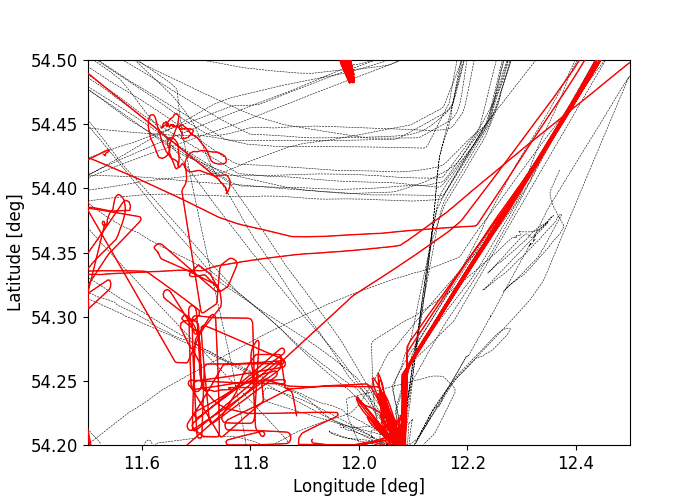}
        \caption{ship type=[0,60), $S_{AT}$=0.1}
    \end{subfigure}%
    % \\
    % \vfill
    \begin{subfigure}{.25\textwidth}
        \centering
        \includegraphics[trim=0 0 40 0, clip, width=1\linewidth]{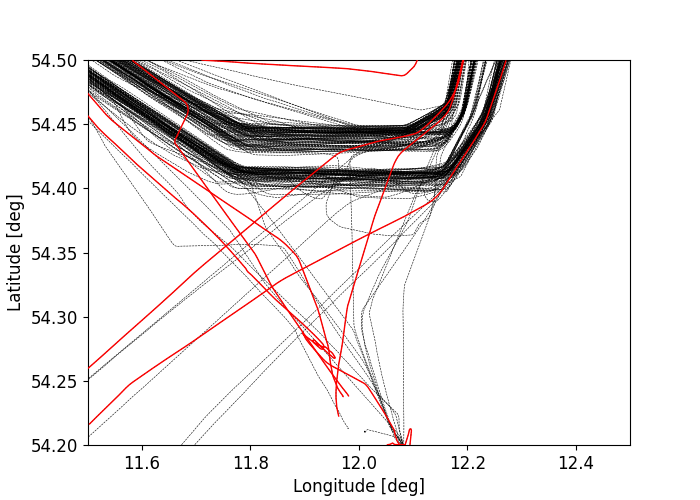}
        \caption{ship type=[60,90), $S_{AT}$=0.3}
    \end{subfigure}%
    \begin{subfigure}{0.25\textwidth}
        \centering
        \includegraphics[trim=0 0 40 0, clip, width=1\linewidth]{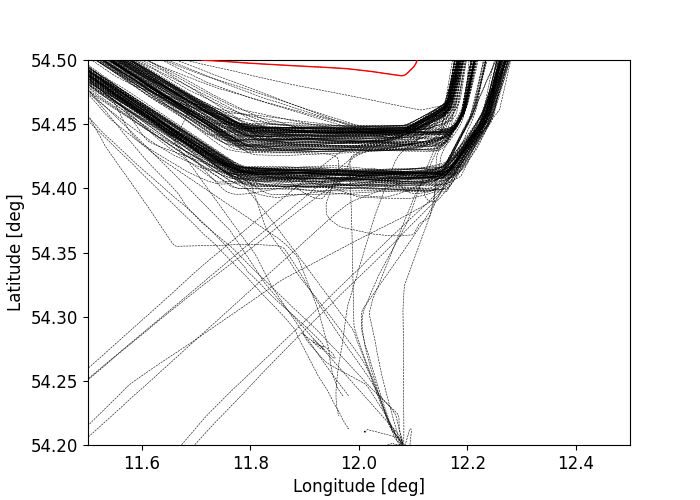}
        \caption{ship type=[60,90), $S_{AT}$=0.1}
    \end{subfigure}%
    \\
    \vfill
    \begin{subfigure}{.25\textwidth}
        \centering
        \includegraphics[trim=0 0 40 0, clip, width=1\linewidth]{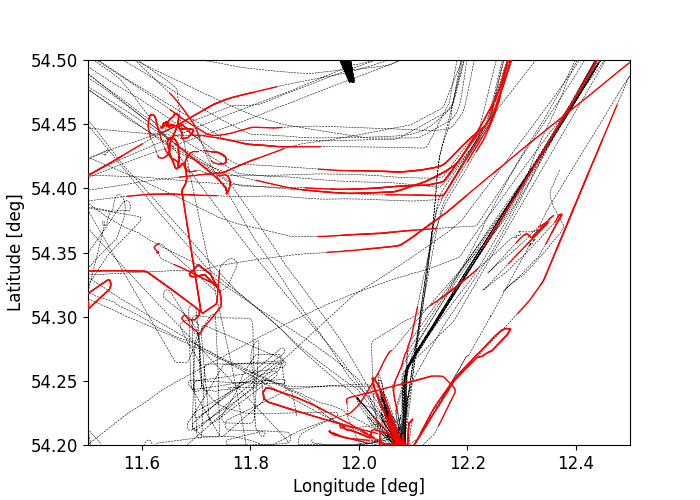}
        \caption{ship type=[0,60), $\Theta_{AT}$=0.7}
    \end{subfigure}%
    \begin{subfigure}{0.25\textwidth}
        \centering
        \includegraphics[trim=0 0 40 0, clip, width=1\linewidth]{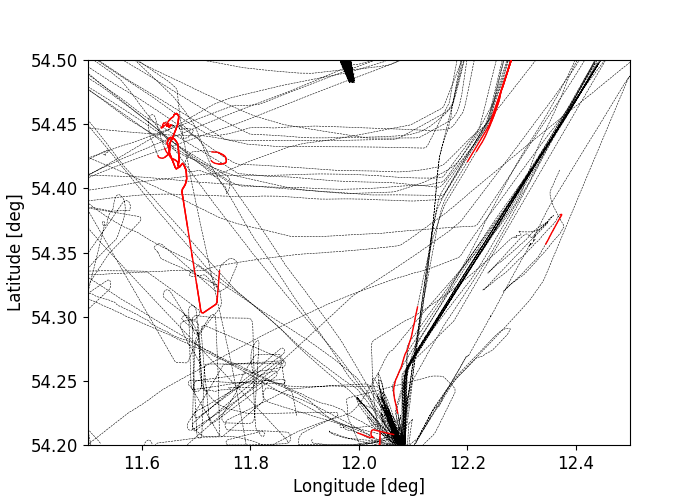}
        \caption{ship type=[0,60), $\Theta_{AT}$=0.4}
    \end{subfigure}%
    \begin{subfigure}{.25\textwidth}
        \centering
        \includegraphics[trim=0 0 40 0, clip, width=1\linewidth]{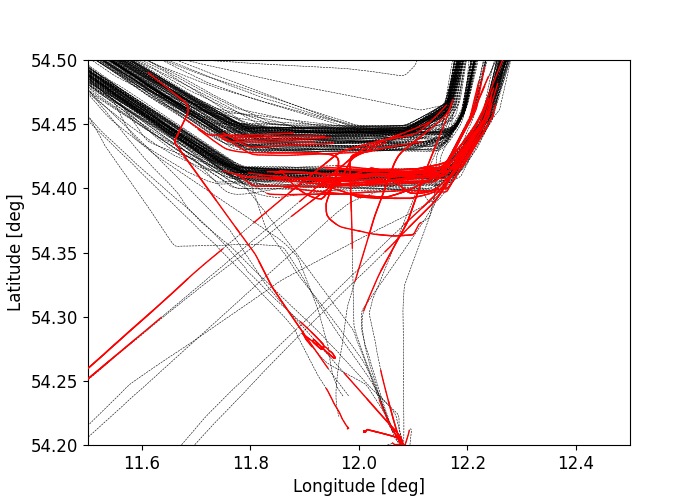}
        \caption{ship type=[60,90), $\Theta_{AT}$=0.7}
    \end{subfigure}%
    \begin{subfigure}{0.25\textwidth}
        \centering
        \includegraphics[trim=0 0 40 0, clip, width=1\linewidth]{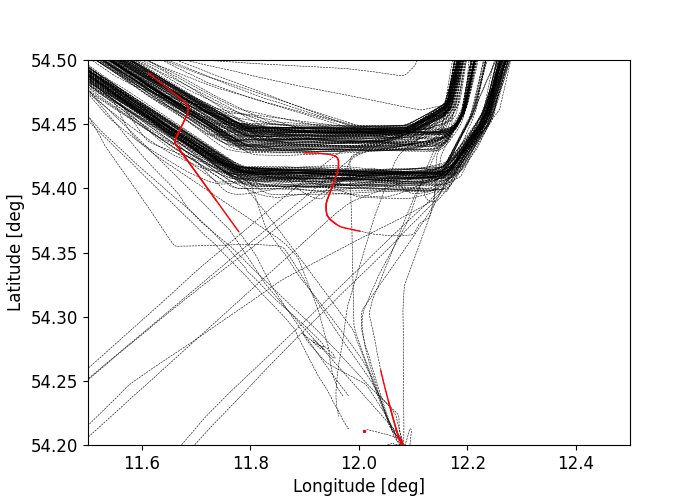}
        \caption{ship type=[60,90), $\Theta_{AT}$=0.4}
    \end{subfigure}%
     \\
    \vfill
    \begin{subfigure}{.25\textwidth}
        \centering
        \includegraphics[trim=0 0 40 0, clip, width=1\linewidth]{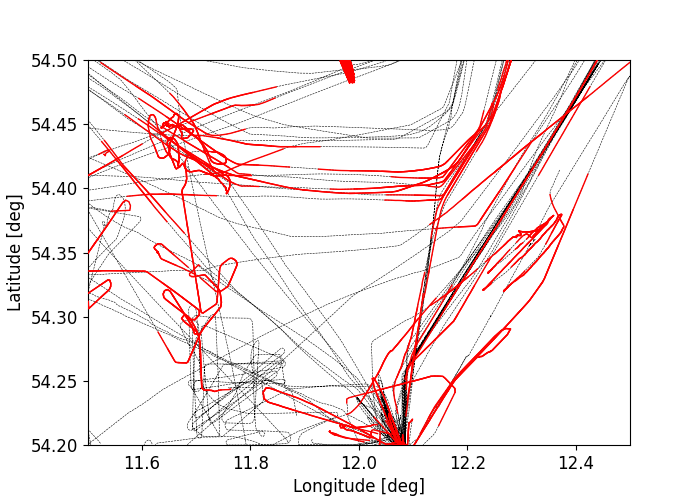}
        \caption{ship type=[0,60), $\Theta_{AT}$=0.7}
    \end{subfigure}%
    \begin{subfigure}{0.25\textwidth}
        \centering
        \includegraphics[trim=0 0 40 0, clip, width=1\linewidth]{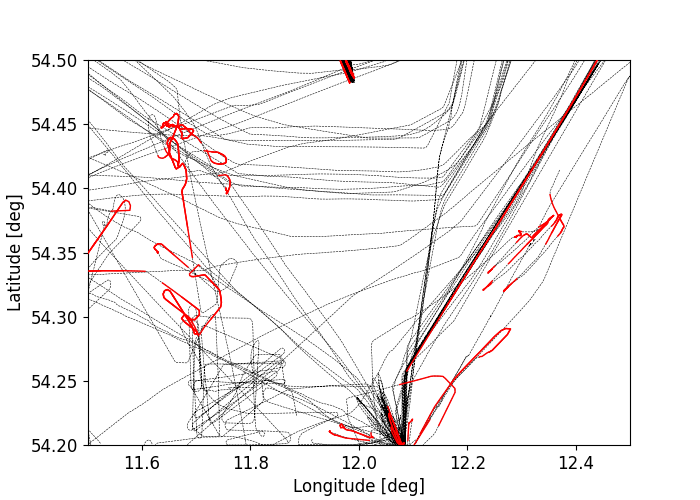}
        \caption{ship type=[0,60), $\Theta_{AT}$=0.4}
    \end{subfigure}%
    \begin{subfigure}{.25\textwidth}
        \centering
        \includegraphics[trim=0 0 40 0, clip, width=1\linewidth]{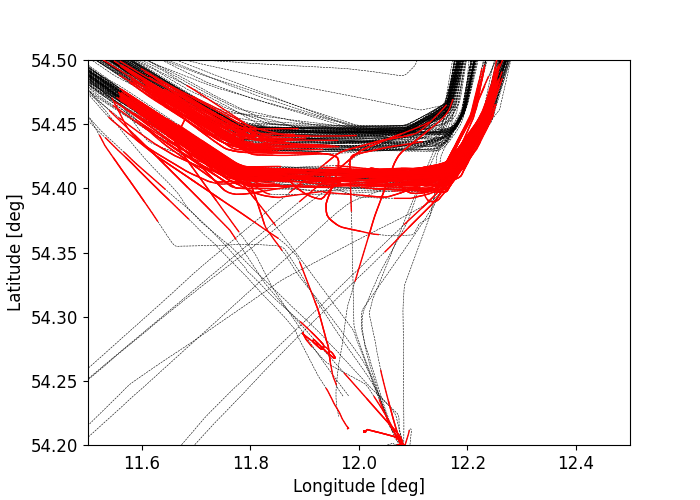}
        \caption{ship type=[60,90), $\Theta_{AT}$=0.7}
    \end{subfigure}%
    \begin{subfigure}{0.25\textwidth}
        \centering
        \includegraphics[trim=0 0 40 0, clip, width=1\linewidth]{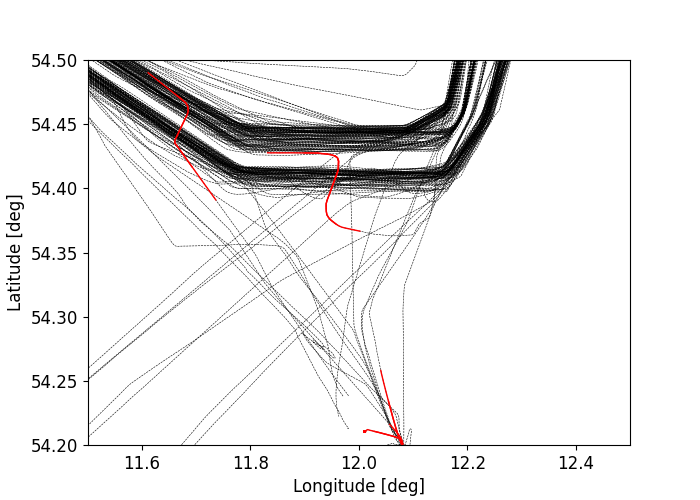}
        \caption{ship type=[60,90), $\Theta_{AT}$=0.4}
    \end{subfigure}
\caption{The figures show anomalous segments of vessels' trajectories (in red) obtained for different thresholds and ship types on dataset $\mathcal{D}_1$. Top, middle, and bottom rows show results for the Similarity, RNN-EDL, and RNN-MC dropout methods, respectively.}  
\label{fig:ano_traj_all_ds1}  
\end{figure*}

% putting percentage between two minipage minimizes gaps
\begin{figure*}[t]
    \centering
        \begin{subfigure}{.25\textwidth}
        \centering
        \includegraphics[trim=0 0 40 0, clip, width=1\linewidth]{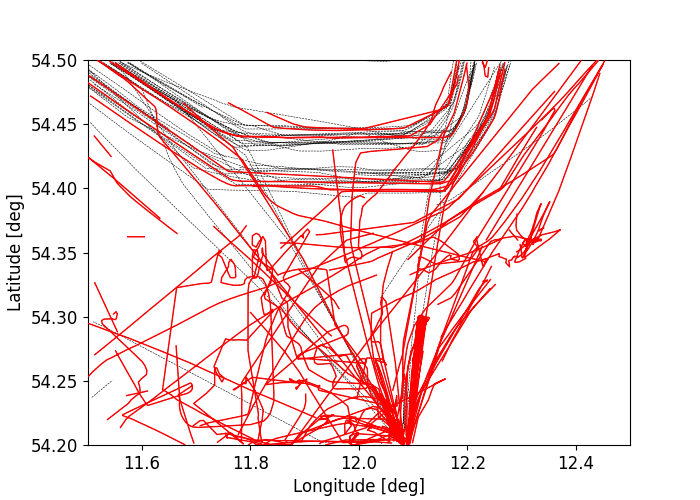}
        \caption{ship type=[0,60), $S_{AT}$=0.3}
    \end{subfigure}%
    \begin{subfigure}{0.25\textwidth}
        \centering
        \includegraphics[trim=0 0 40 0, clip, width=1\linewidth]{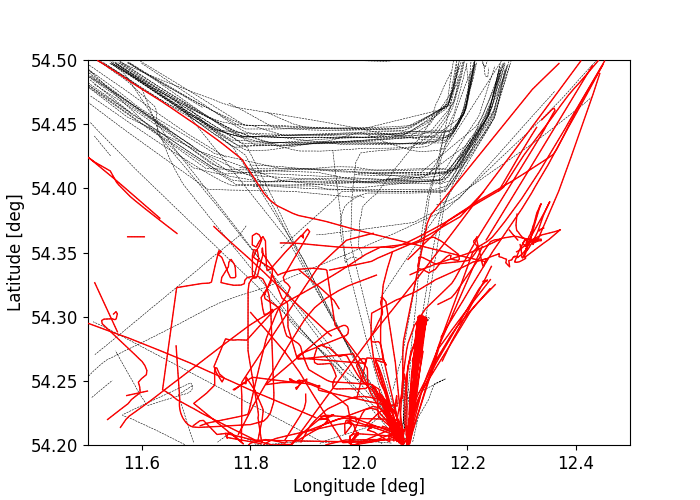}
        \caption{ship type=[0,60), $S_{AT}$=0.1}
    \end{subfigure}%
    % \\
    % \vfill
    \begin{subfigure}{.25\textwidth}
        \centering
        \includegraphics[trim=0 0 40 0, clip, width=1\linewidth]{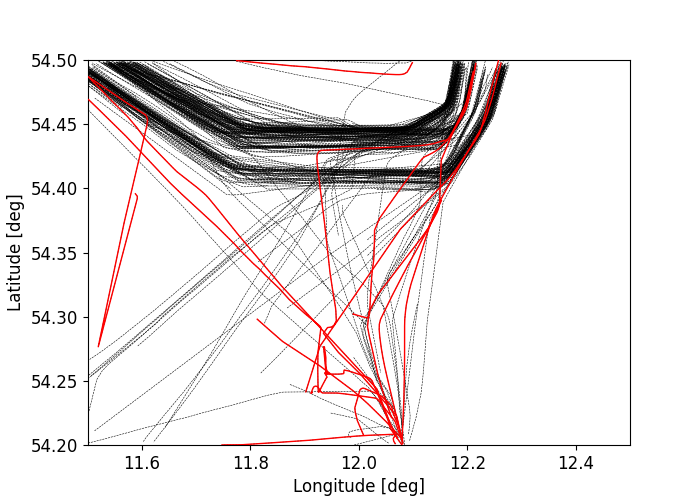}
        \caption{ship type=[60,90), $S_{AT}$=0.3}
    \end{subfigure}%
    \begin{subfigure}{0.25\textwidth}
        \centering
        \includegraphics[trim=0 0 40 0, clip, width=1\linewidth]{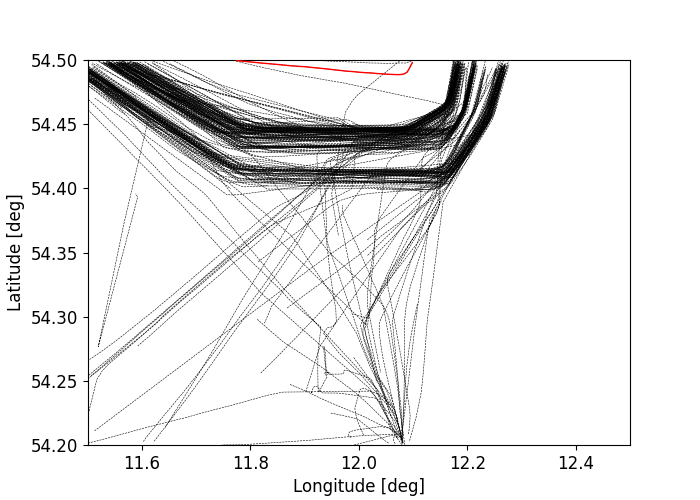}
        \caption{ship type=[60,90), $S_{AT}$=0.1}
    \end{subfigure}%
    \\
    \vfill
    \begin{subfigure}{.25\textwidth}
        \centering
        \includegraphics[trim=0 0 40 0, clip, width=1\linewidth]{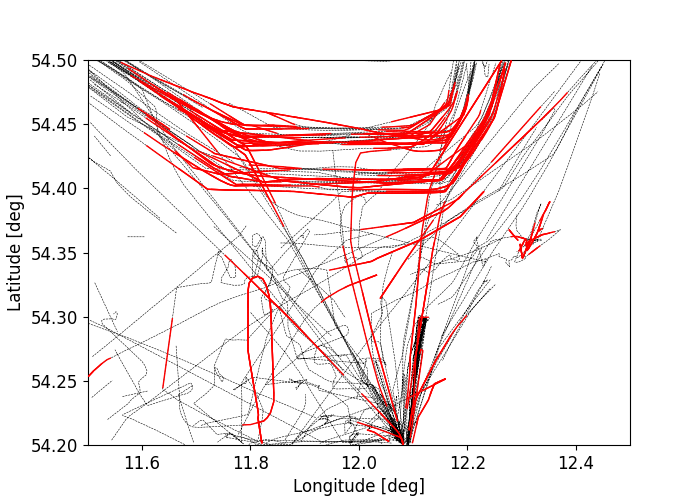}
        \caption{ship type=[0,60), $\Theta_{AT}$=0.7}
    \end{subfigure}%
    \begin{subfigure}{0.25\textwidth}
        \centering
        \includegraphics[trim=0 0 40 0, clip, width=1\linewidth]{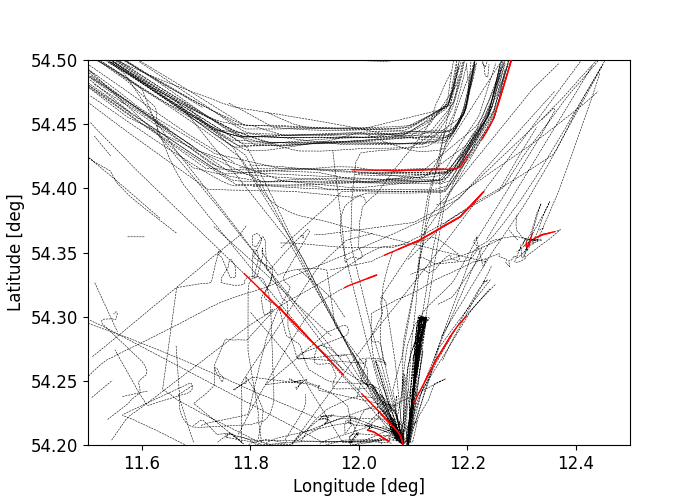}
        \caption{ship type=[0,60), $\Theta_{AT}$=0.4}
    \end{subfigure}%
    \begin{subfigure}{.25\textwidth}
        \centering
        \includegraphics[trim=0 0 40 0, clip, width=1\linewidth]{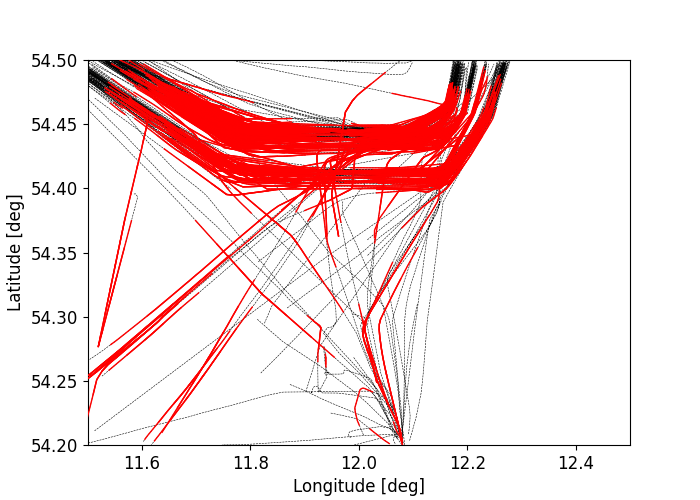}
        \caption{ship type=[60,90), $\Theta_{AT}$=0.7}
    \end{subfigure}%
    \begin{subfigure}{0.25\textwidth}
        \centering
        \includegraphics[trim=0 0 40 0, clip, width=1\linewidth]{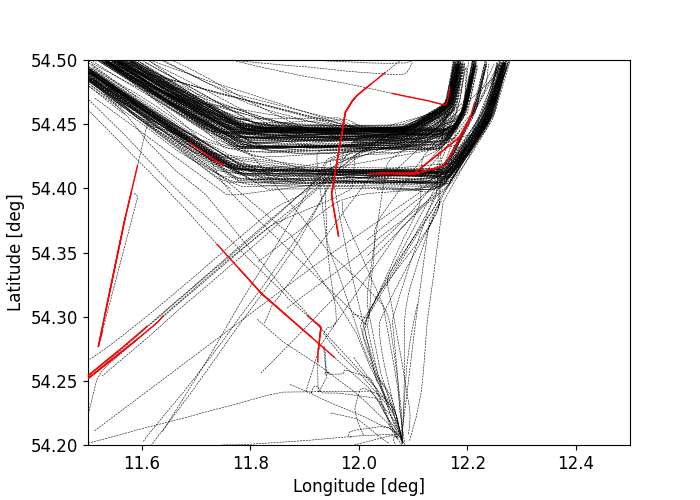}
        \caption{ship type=[60,90), $\Theta_{AT}$=0.4}
    \end{subfigure}%
     \\
    \vfill
    \begin{subfigure}{.25\textwidth}
        \centering
        \includegraphics[trim=0 0 40 0, clip, width=1\linewidth]{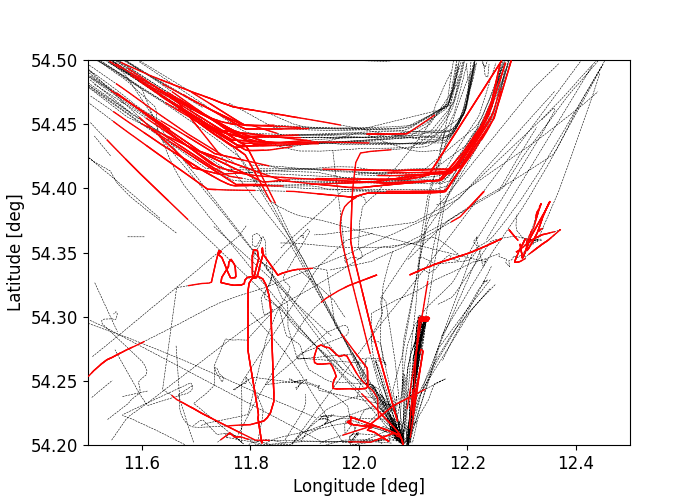}
        \caption{ship type=[0,60), $\Theta_{AT}$=0.7}
    \end{subfigure}%
    \begin{subfigure}{0.25\textwidth}
        \centering
        \includegraphics[trim=0 0 40 0, clip, width=1\linewidth]{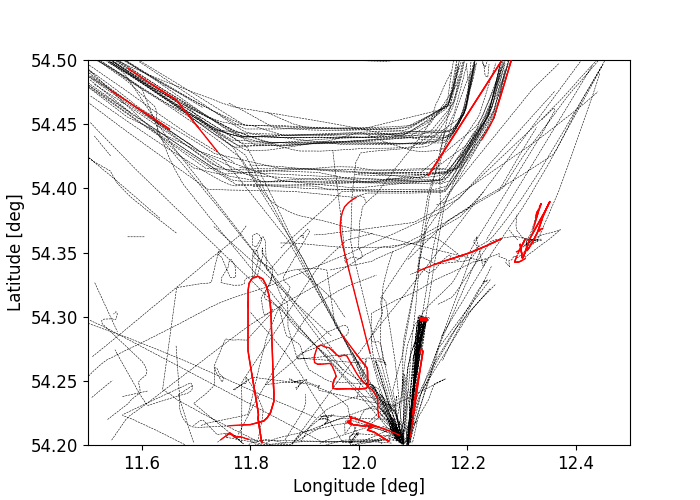}
        \caption{ship type=[0,60), $\Theta_{AT}$=0.4}
    \end{subfigure}%
    \begin{subfigure}{.25\textwidth}
        \centering
        \includegraphics[trim=0 0 40 0, clip, width=1\linewidth]{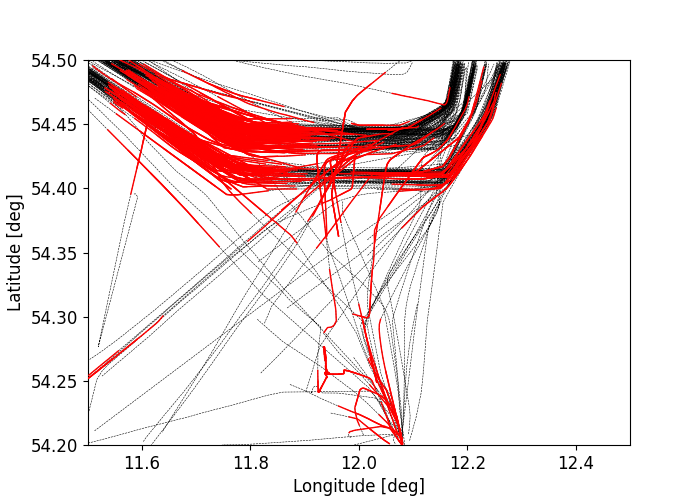}
        \caption{ship type=[60,90), $\Theta_{AT}$=0.7}
    \end{subfigure}%
    \begin{subfigure}{0.25\textwidth}
        \centering
        \includegraphics[trim=0 0 40 0, clip, width=1\linewidth]{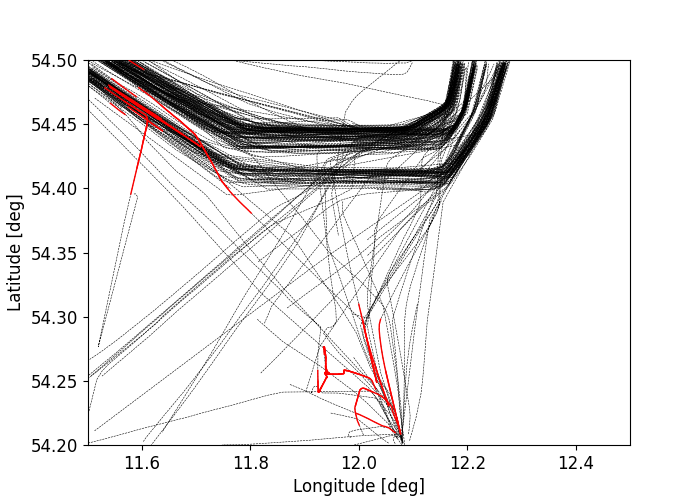}
        \caption{ship type=[60,90), $\Theta_{AT}$=0.4}
    \end{subfigure}
\caption{The figures show anomalous segments of vessels' trajectories (in red) obtained for different thresholds and ship types on dataset $\mathcal{D}_2$. Top, middle, and bottom rows show results for the Similarity, RNN-EDL, and RNN-MC dropout methods, respectively.}  
\label{fig:ano_traj_all_ds3}  
\end{figure*}

% \begin{figure*}[t]
%     \centering
%     \begin{subfigure}{.49\textwidth}
%         \centering
%         \includegraphics[width=1\linewidth]{figures/anoall_graph_True_ship_60_90_ds3_unc_3.png}
%         \caption{ship type $=[60,90), \Theta_{AT}=0.7$}
%     \end{subfigure}
%     \begin{subfigure}{0.49\textwidth}
%         \centering
%         \includegraphics[width=1\linewidth]{figures/anoall_graph_True_ship_0_60_ds3_unc_3.png}
%         \caption{ship type $=[0,60), \Theta_{AT}=0.7$}
%     \end{subfigure}
%     \\
%     \begin{subfigure}{.49\textwidth}
%         \centering
%         \includegraphics[width=1\linewidth]{figures/anoall_graph_True_ship_60_90_ds3_unc_6.png}
%         \caption{ship type $=[60,90), \Theta_{AT}=0.4$}
%     \end{subfigure}
%     \begin{subfigure}{0.49\textwidth}
%         \centering
%         \includegraphics[width=1\linewidth]{figures/anoall_graph_True_ship_0_60_ds3_unc_6.png}
%         \caption{ship type $=[0,60), \Theta_{AT}=0.4$}
%     \end{subfigure}
% \caption{The figures show anomalous segments of vessels' trajectories obtained using prediction uncertainty from 30 continuous sequences for different uncertainty thresholds and ship types on $\mathcal{D}_2$.}  
% \label{fig:ano_traj_all_ds2}  
% \end{figure*}

\begin{figure*}[ht!]
\centering
\includegraphics[trim=60 0 60 0, clip, width=1\textwidth]{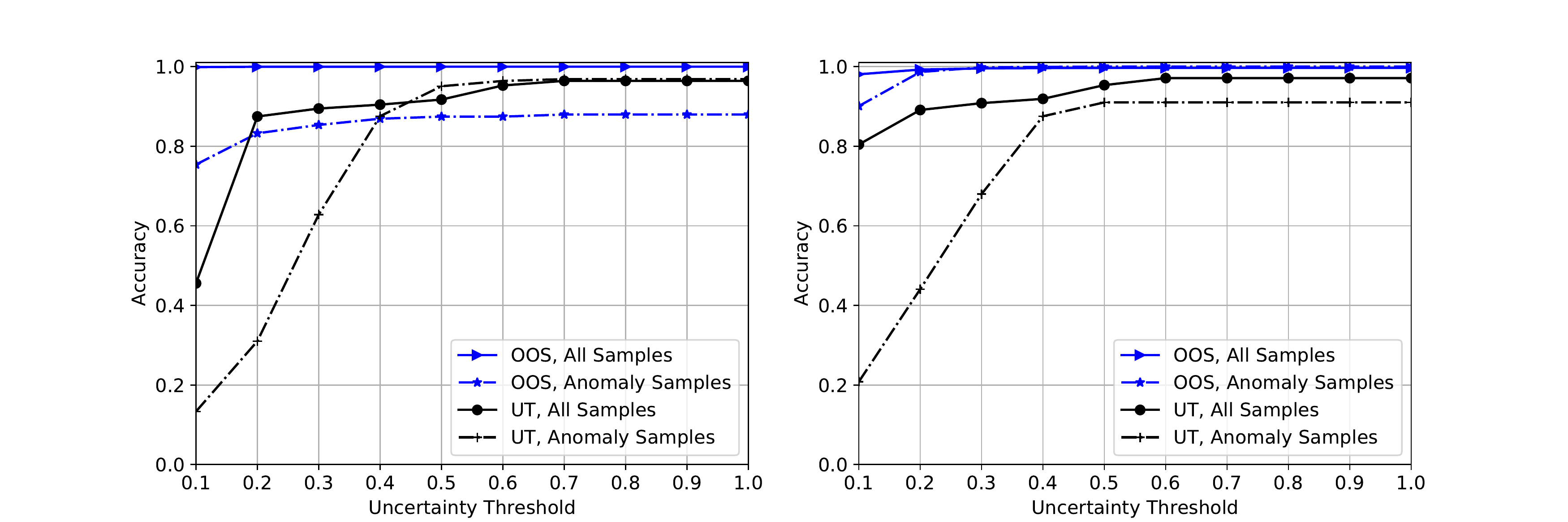}
%\vspace{-0.2 cm}
 \caption{Left and right figures show the accuracy w.r.t. uncertainty threshold by the AIS OOS classifier and the unusual turn (UT) classifier evaluating all samples and only the anomalous samples in test datasets $\mathcal{D}_1$ and $\mathcal{D}_2$, respectively.}
% \vspace{-0.4cm}
\label{fig:acc_unc_all}
\end{figure*}

\begin{figure*}[t]
    \centering
    \begin{minipage}{.48\textwidth}
        \centering
        \includegraphics[trim=5 0 5 10, clip,width=1\linewidth]{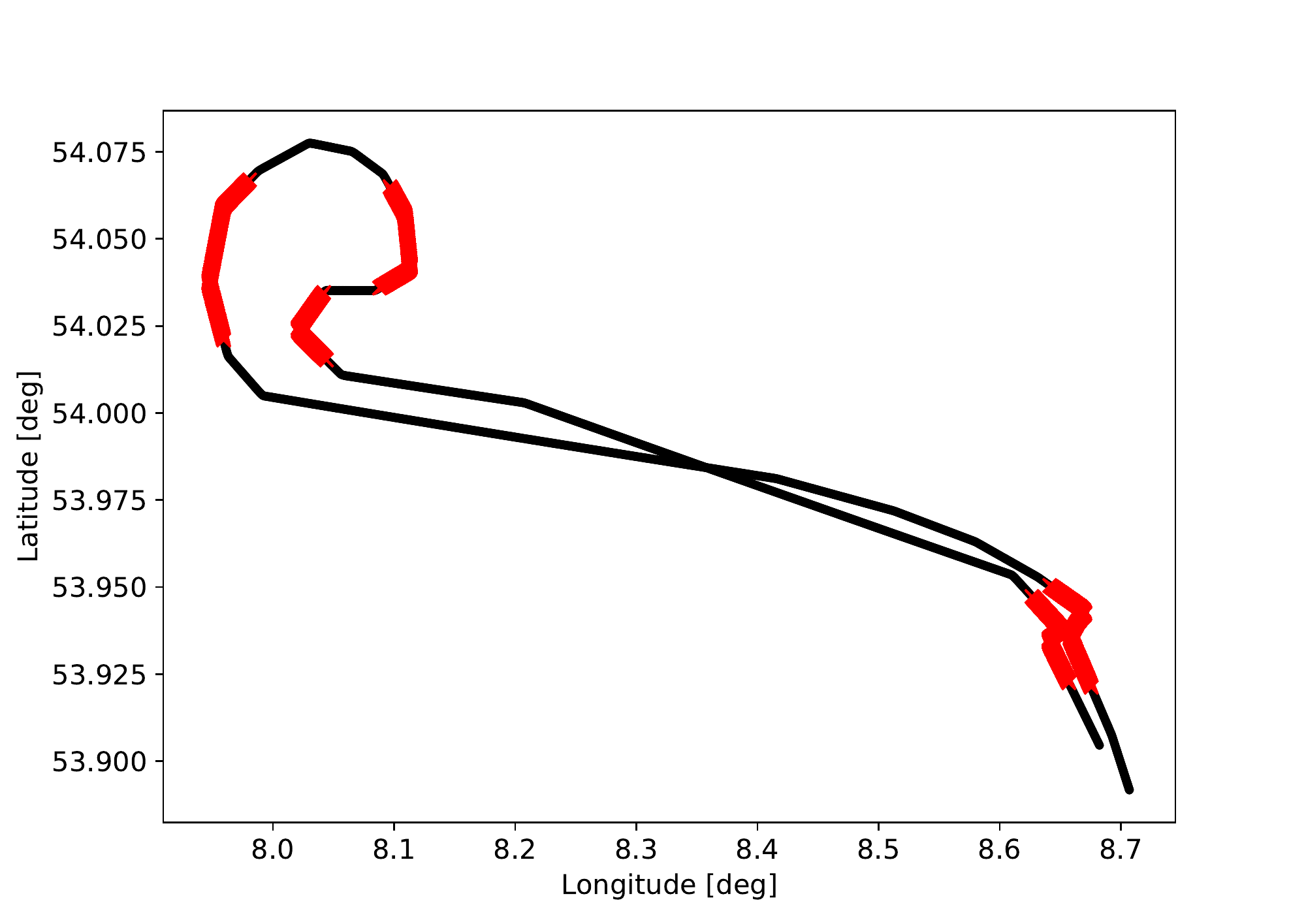}
    \end{minipage}
    \begin{minipage}{0.48\textwidth}
        \centering
        \includegraphics[trim=5 0 5 10, clip, width=1\linewidth]{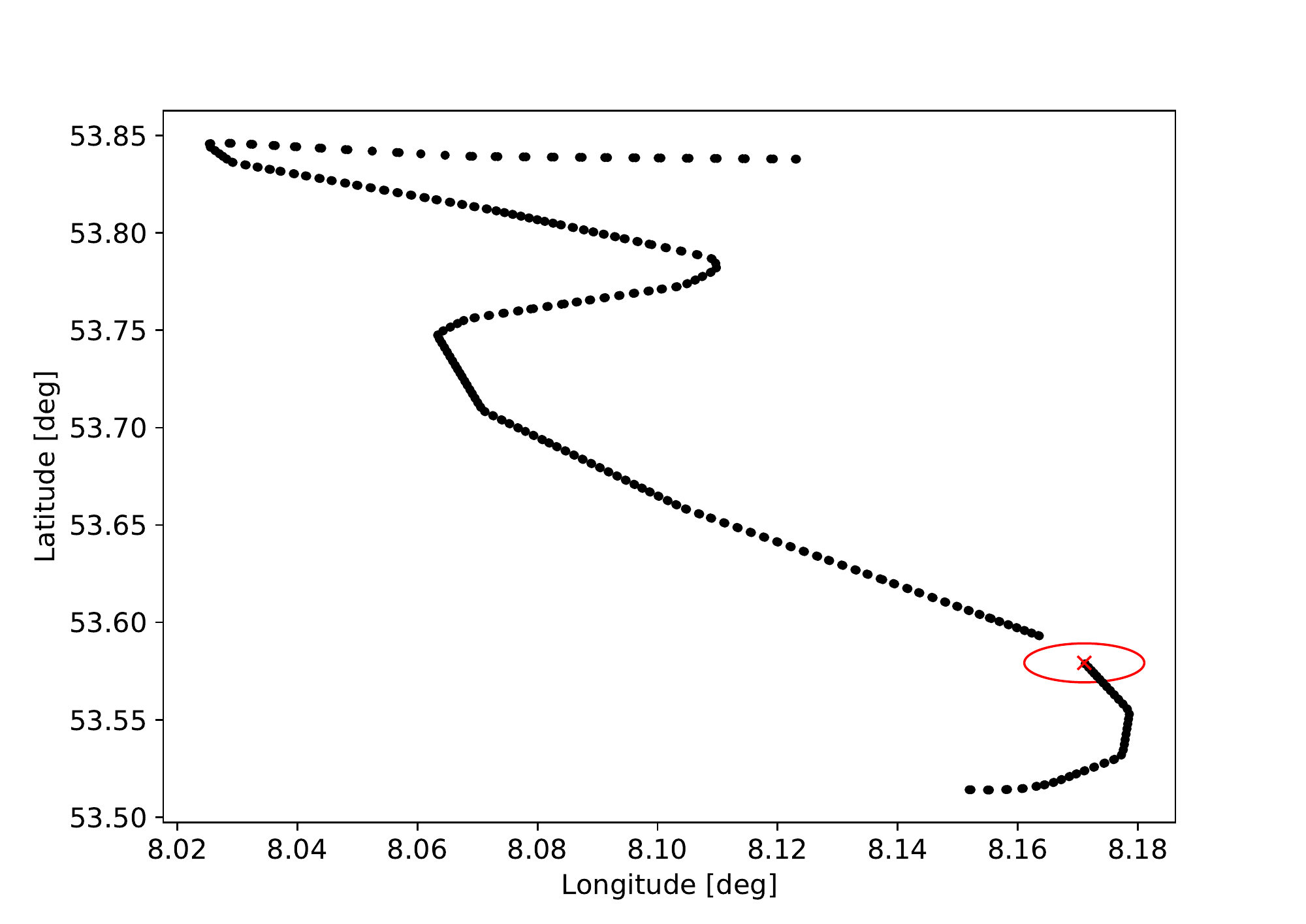}
    \end{minipage}
\caption{The left plot shows the unusual turn ($> 30\degree$) segments (red) detected  on a test vessel's trajectory (black). The right plot shows the OOS (red) detected  on another test vessel moving from bottom to up direction (black).}  
\label{fig:ano_test_vessel}  
\end{figure*}

\subsection{GTRA Evaluation}
Before showing the effectiveness of the GTRA scheme in trajectory prediction and uncertainty estimation, we present the results on how a graph can represent vessel movement patterns using dataset $\mathcal{D}_1$. Here, we consider only passenger, cargo and tanker vessels, as they follow well-defined trajectories. Fig. \ref{fig:graph_all} (left) depicts the original graph in black that is formed by data plotted in blue. Notice that the nodes are the centred at waypoints (red dots) with high density. The bidirectional edges very well represent vessel motion patterns. However, we can also see that the GTR algorithm places a node around (12.1\degree, 54.3\degree) without any further edge connection, misrepresenting underlying vessel movements. 

Note that, since the parameters of RDP and DBSCAN are sensitive to the dataset in question, the computed graphs might be inadequate to represent all of the important sea lanes. This is particularly true in scenarios where the AIS-range is limited. There may be trajectories that are recorded in-between out-of-range source(s) or destination(s). Therefore, an expert knowledge that might improve the traffic representation, for example, by manually adding a node/edge to account for such limitation.
Thus, as an offline process we refined the graph by deleting the node and connecting two edges, as shown in Fig. \ref{fig:graph_all} (right). The latter additionally shows the association of data points to one of the edges (clusters) in a color-coded fashion where a so-called outlier edge is represented in grey. Notice that the majority of movements show acceptable assignment to edges. Only some segments of trajectories have been wrongly (though subjective) assigned owing to their complex maneuvers.

\textbf{Impact of GTRA on prediction accuracy}:  Trajectory representation and clustering has huge impact on the accuracy of trajectory prediction  as well as on the corresponding uncertainty estimation. Fig. \ref{fig:cdf_dist_error} (left) illustrates this observation using a cumulative distribution/density function (CDF), which is obtained on the Euclidean distances (or positional errors) between ground truth and predicted positions on  trajectories of ships in the dataset $\mathcal{D}_1$. Aside from the superior performance of the graph-based prediction model, we can also notice that it is challenging to predict trajectories of ships in range [0, 60), as compared to passenger, cargo and tanker vessels ([60, 90)). More importantly, the RNN-EDL model without graph-based clustering predicts more than 80\% of target positions with deviation by at least 30 km from the ground-truth data. The reason is that the model fails to fit targets with the Gaussian distribution, as the maritime data is dispersed. The clustered data, on the other hand, simplifies the model's ability to learn the motion pattern and fit the distribution. 
To further investigate the impact of the EDL layer, the latter was removed from the RNN-EDL model. The trained RNN model is then used for the prediction of trajectories, and to plot CDFs of positional errors in Fig. \ref{fig:cdf_dist_error} (right). We observe that the prediction error significantly decreased for the RNN model without GTRA. Nevertheless, RNN-EDL with graph in Fig. \ref{fig:cdf_dist_error} (left) outperforms others. In the next subsection, we show results of only the RNN-EDL model with GTRA. 

\subsection{Anomalous Trajectories Detection}
We evaluate three trajectory segments: a \emph{normal} track and two anomalous tracks in the \emph{left} and \emph{right} to get an insight on the trajectory prediction in Fig. \ref{fig:pred_unc_rg_left_normal_right}(left). Note that the left and right tracks are synthetic data based on recorded data from a real-world (or normal) track in the middle by changing the longitudes and latitudes. We can see from Fig. \ref{fig:pred_unc_rg_left_normal_right} (left) that the ground truth data closely follows the predicted ones. Given an input sequence, the model is able to correctly forecast the vessel's position at a future time step not only for the normal track but also for the unseen (left and right) trajectories, which would not be the case with the other models. %Note that when we tried to predict the anomalous trajectories with the regression model without the \emph{edge} feature, all three predicted  trajectories followed the same normal segment (the plot is omitted here). 

Fig.  \ref{fig:pred_unc_rg_left_normal_right} (right) depicts the epistemic uncertainties on successive predictions for the three segments and over the $lon, lat, cog, sog$ data features, respectively. We observe that the left and right tracks show an abrupt transition (or jump) in epistemic uncertainties with respect to the normal for all features. The change in epistemic uncertainty for the anomalous tracks are expected for the out-of-order data distribution. Moreover, as the anomalous segments deviate from the normal, their data might get assigned to different edge clusters, including an outlier edge, and produce jumps in uncertainties. For this reason, aleatoric uncertainties can also show abrupt transitions for heterogeneous maneuvering. 
%As expected, the aleatoric uncertainty values are higher (lower) than the epistemic for $lon$ and $lat$ ($cog$ and $sog$) features. This is because it represents the data noise, and we can see in Fig. \ref{fig:rostock_gedsar} that the data ($lon, lat$) of AIS trajectories with the same cluster (i.e., the north-south edge in Fig. \ref{fig:graph_all}) get noisier around these coordinates. 
Also, note that despite having the same $cog$ and $sog$ values in all three segments, the epistemic uncertainties in $cog$ and $sog$ clearly show the transitions for the left and right tracks. This shows that the model learnt the correlation among features well. It is also important to mention that when we trained the RNN-EDL regressor model with only the Rostock-Gedsar dataset (see trajectories in Fig. \ref{fig:rostock_gedsar}), the relative change in the uncertainties is even higher (not shown for sake of brevity), as it is easier for the model to learn the data distribution of  trajectories between two ports accurately. However, we are interested in developing a single global model to detect all anomalous trajectories in the ROI. 

Next, Figures \ref{fig:ano_traj_all_ds1} and \ref{fig:ano_traj_all_ds3} show all anomalous segments of each track in red for $\mathcal{D}_1$ and $\mathcal{D}_2$, respectively. The results are obtained for two sets of vessels, two detection thresholds using three different models, the Similarity, RNN-EDL and RNN-MC dropout, and shown on the top, middle and bottom rows, respectively. The similarity thresholds are $S_{AT}=\{0.3, 0.1\}$. The thresholds $\Theta_{AT}=\{0.7, 0.4\}$, and segment size $N=30$ are used for both the RNN-EDL and RNN-MC dropout models to detect anomalous trajectories using Eq. \eqref{eqn:at_def}. The epistemic (aleatoric) uncertainty is used for anomalous trajectory detection by the RNN-EDL (RNN-MC dropout) using Eq. \eqref{eqn:at_def}. It is important to mention that the epistemic uncertainty in the MC dropout resulted in detecting many more anomalous trajectories at these thresholds. 
%. In contrast,
Unlike the EDL or MC dropout, the Similarity model uses complete track or a larger segment to compute waypoints and the similarity score for anomalous trajectory detection. 
Obviously, decreasing the threshold decreases the number of detected anomalous segments, as the threshold is an upper bound in Eq. \eqref{eqn:at_def} and a lower threshold value only detects larger transitions. This is true for the Similarity model as well, since two similar tracks have a higher similarity score. As expected, the Similarity model detects ATs which are non-overlapping and further away from the commonly traversed routes (formed by graph edges). In contrast, DL models are able to detect anomalous segments that show unusual maneuvering and lie across edges. Thus, the similarity and DL models are complementing each other in finding ATs outside and within clusters. Both DL models perform well at lower threshold $\Theta_{AT}=0.4$. Additionally, we make the following common observations about the three models. The passenger, cargo and tanker ships' trajectories with type [60,90) in the right two plots show lower number of anomalous segments than other ship types shown in the left two plots at lower threshold. We can say that the models are able to detect some unusual maneuvering while missing out on some tracks that seem to show anomalous behavior. At the same time, some detected anomaly segments do not seem to be anomalous after a careful look at their tracks. Nevertheless, we observe that having a common threshold for all ships is not the best solution. Instead, we can set a lower threshold for passenger, cargo, and tanker vessels, and a higher threshold for other vessel types. Another possible solution could be to find uncertainty thresholds of normal tracks, though computationally intensive and subjective in nature, in smaller grids of the ROI and compare them with the estimated uncertainty on any trajectory under evaluation.  

\subsection{Unusual Turn and OOS Detection}
For an unusual turn detection, we choose a threshold $\Theta_{UT} = 30\degree$, considering that larger vessels generally do not take more than $30\degree$ of turn within a short interval $T\times \tau = 3$ minutes, unless constrained by their geographical locations (which is not considered here). Thus, we evaluate the model accuracy only for passenger, cargo and tanker vessels. On the other hand, we set a threshold $\Theta_{OOS} = 3$ minutes for the OOS anomaly detection (Eq. \ref{eqn:oos_def}). 

In Fig. \ref{fig:acc_unc_all} we show the accuracy of the EDL classifiers against the uncertainty threshold for all samples as well as for only anomalous samples in the datasets. We accept the classification of a sample into a class (normal or anomaly) if it has the highest probability and the prediction uncertainty is below a threshold $u_{th}$. In other words, the sample's classification by the model is rejected if the epistemic uncertainty is above a threshold. The following observations are made. Firstly, increasing the threshold improves the accuracy, which suggests that the models fits well to the Dirichlet categorical distribution. Secondly, the UT model classifies most of the samples with higher uncertainty ($> 0.2$) as compared to the OOS classifier ($< 0.1$), and it achieves 90\% accuracy if the uncertainty threshold $u_{th} = 0.4$. Thirdly, the UT classifier not only misclassifies anomalous samples as normal, but also other way round, which can be seen from the accuracy obtained on all samples and only the anomalous samples. Finally, the OOS classifier achieves an overall accuracy of $\approx$ 99\% on all samples for uncertainty thresholds greater than 0.1. 
The accuracy of the EDL models for UT and OOS detection on only anomalous samples are lower at uncertainty thresholds below 0.4, compared to the accuracy computed on all samples. The reason is that the models produce larger uncertainties on their predictions. This suggests that the models find it relatively hard to learn anomalous samples than the normal samples. %Nevertheless, we also see that models accuracy  that the min $\mathcal{D}_1$ is relatively lower at $82$\%. The reason could be that the model was trained with a smaller ($\approx 500$) number of normal and OOS data samples, as $\mathcal{D}_1$ has only 0.02\% OOS samples. 

%However, when we evaluate only anomaly samples, we find that the accuracy dips to 82\% on dataset $\mathbf{D}_1$, as the EDL model was trained with smaller ($\~ 500$) number of data samples per class than the EDL model trained on $\mathbf{D}_2$ with 6\% ($\~ 4500$) OOS anomalous samples. Nevertheless, the performance of both the models on the tested samples of both datasets are acceptable.

Unlike the RNN-EDL regressor model, the trained EDL classifiers on ROI data could be used to detect UT and OOS anomalies on AIS data beyond ROI. We, therefore, evaluate the EDL classifiers on two additional scenarios: one for the dump cleaning (UT), and other for the AIS blackout (OOS) exercises that was performed by a test vessel in a validation campaign of a project in July 2020 in the Bremerhaven region in Germany. The uncertainty threshold $u_{th}$ is set to 0.4. Fig. \ref{fig:ano_test_vessel} (left) shows all the unusual turn segments detected, and the right plot shows the detected AIS blackout event in red in the test vessel's trajectories. We observe that an unusual turn segment on the upper left region is undetected in the left plot. Nevertheless, the classifiers are able to detect all other UT and OOS anomalies well.

\section{Concluding Remarks} \label{sec:conclusion}  
\par We proposed a novel graph-based trajectory representation and association (GTRA) method, and deep learning models to detect maritime anomalies using AIS data. We showed that the traffic representation and association are key for learning the AIS traffic patterns and the detection of anomalous trajectories by the DL-based regression models. Furthermore, we evaluated three different anomalies: anomalous trajectory using the EDL and MC dropout regressor models, and the loss of AIS signal and unusual turn by EDL classifiers using the uncertainties in prediction. The anomalous tracks are detected by evaluating jumps in uncertainties on trajectory segment prediction. Similarly, the unusual turn anomaly and the loss of AIS signal are detected by taking into account the class probabilities and prediction uncertainty. In summary, we make the following observations based on the proposed models and analysis presented in the paper.
\begin{itemize}
    \item The GTRA is essential for the RNN-EDL time-series prediction model in order to learn distribution of maritime trajectories. In contrast to an RNN model without the EDL layer, the prediction performance of the RNN-EDL model decreases significantly when trained without the graph-based clustered data.
    \item The epistemic and aleatoric uncertainties obtained on  prediction by the DL regressor models are very useful in anomalous trajectory detection. These models together with a similarity model can detect anomalous trajectories within and outside sea-lanes. While DL classifier models can be useful in detecting unusual turn and the loss of AIS signal.
    \item Typically, a lower detection threshold for passenger, cargo and tankers vessels, and higher for other ship types can be set to detect anomalous tracks. Additionally, large vessels taking usually fixed routes show lower number of anomalous track segments than other vessels, for example, fishing. 
\end{itemize}

This work opens deeper research activities within the field of leveraging deep learning uncertainties in maritime anomaly detection. Moreover, the work might be useful in detecting anomalies in other domains of transportation. Lastly, in the absence of  reference anomalous routes, expert knowledge will always be required before taking any action upon detection of anomalies.
%%\vspace{-0.3cm}
\section*{Acknowledgment}
We would like to thank Dr. Frank Heymann and Juan C. Machuca for providing valuable inputs that shaped the paper. This work has been partially funded by the German federal ministry for economic affairs and energy project \textit{IntelliMar}, and the German federal ministry of education and research project \textit{European Maritime Safety III}.

% \pagebreak
% \newpage
\bibliographystyle{IEEEtran}
\bibliography{final_manuscript}

\begin{IEEEbiography} 
[{\includegraphics[width=1in,height=1.25in,clip,keepaspectratio]{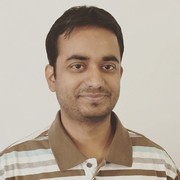}}]{Sandeep Kumar Singh} (SM'15, M'20)  is a postdoctoral researcher in the department of Electrical and Computer Engineering at the University of California, Davis, USA. He received his PhD in Computer and Network Engineering from the Technical University of Braunschweig, Germany in 2019, and M.S. degree in Electrical Engineering from the Indian Institute of Technology, Madras, India in 2014. He was a research scientist at the Institute of Communication and Navigation, at the German Aerospace Center, Germany from 2019 to 2021. He was a recipient of German Scholarship DAAD in 2011-12 to pursue research at the Technical University of Berlin, Germany. His research interests include communication and navigation, data center networking, quantum communication, and machine learning.
\end{IEEEbiography}
 
\begin{IEEEbiography}
[{\includegraphics[width=1in,height=1.25in,clip,keepaspectratio]{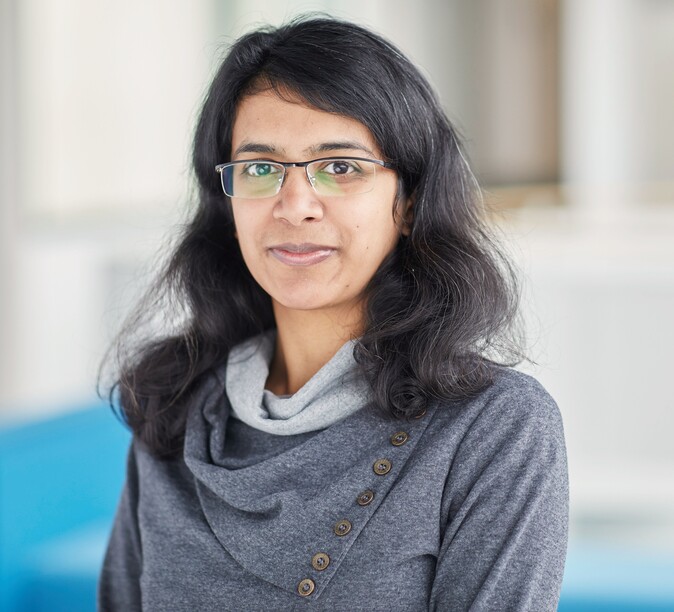}}]{Jaya Shradha Fowdur} is currently working towards her PhD in Computer Science with the Institute of Communications and Navigation at the German Aerospace Centre and the University of G\"ottingen, Germany since 2018. She received her Master's degree in Applied Computer Science with specialisation in Computational Neuroscience from the University of G\"ottingen in 2017. Her research interests include clustering, machine learning, target tracking, and AIS- and radar- data processing in the maritime domain.
\end{IEEEbiography}

\begin{IEEEbiography}[{\includegraphics[width=1in,height=1.25in,clip,keepaspectratio]{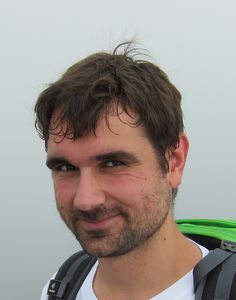}}]{Jakob Gawlikowski} received the bachelor's and master's degree in mathematics from the Technical University of Munich, Munich, Germany in 2015 and 2019. 
\par
He is currently a researcher at the "Institute of Data Science", German Aerospace Center (DLR), Jena, Germany and working towards the
Ph.D. degree with the Technical University of Munich, Munich, Germany, and the German Aerospace Center in Munich, Munich, Germany.
\par
His research interests are related to multi-modal machine learning, uncertainty quantification and robustness in deep learning models with a special focus on applications in the field of remote sensing. 
\end{IEEEbiography}

\begin{IEEEbiography}
[{\includegraphics[width=1in,height=1.25in,clip,keepaspectratio]{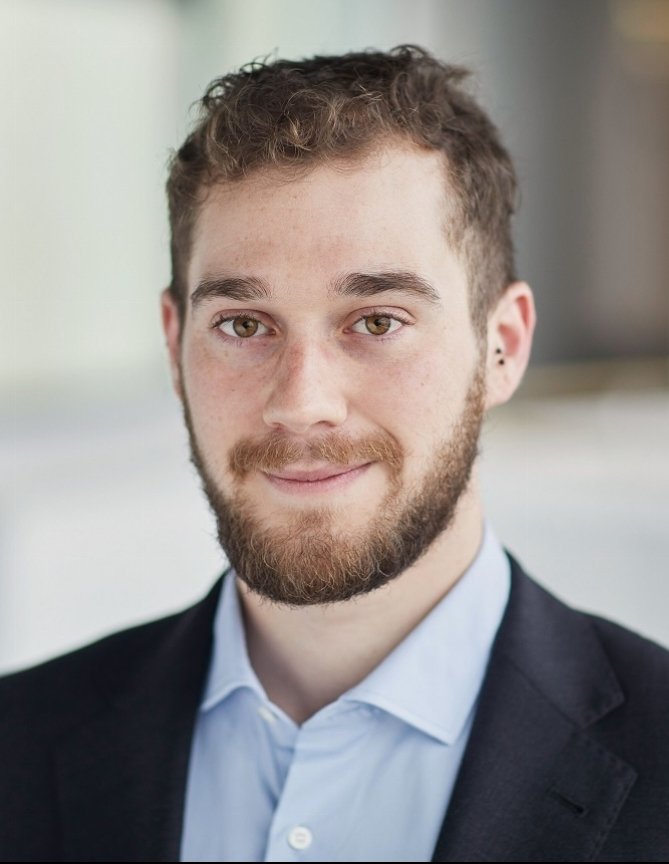}}] {Daniel Medina} (SM'16, M'20) his B.S. in Electrical Engineering from the University of Malaga and his M.S. in Computer Science from University Carlos III Madrid in 2014 and 2016 respectively. He is at the verge of completing his PhD on robust filtering and satellite-based precise localization and attitude estimation. Since 2016, he has been a research fellow at the Institute of Communications and Navigation of the German Aerospace Center. He is currently the Vice-Chair for the IEEE ITSS Germany Chapter. His primary areas of interest include signal processing, estimation theory, nonlinear and robust  filtering with applications to tracking, navigation and remote sensing.
\end{IEEEbiography}

\end{document}